\PassOptionsToPackage{table}{xcolor}
\PassOptionsToPackage{numbers}{natbib}

\documentclass{article}
\usepackage[utf8]{inputenc}
\usepackage[preprint]{colm}
\setcitestyle{numbers,square,citesep={;}}
\usepackage{microtype}
\usepackage{graphicx}
\usepackage{booktabs} 
\usepackage{comment}
\usepackage{arydshln}
\usepackage{makecell}
\usepackage{hyperref}
\usepackage{mathtools}
\usepackage{color}
\usepackage{listings}
\usepackage{xcolor}

\usepackage{graphicx}
\usepackage{CJKutf8}
\usepackage{url}
\usepackage{multicol,multirow}
\usepackage{makecell}
\usepackage{algorithmic}
\usepackage{algorithm}
\usepackage{booktabs}
\usepackage{textcomp}

\usepackage{bm}
\usepackage{parskip}
\usepackage[T1]{fontenc}    
\usepackage{booktabs}       
\usepackage{amsmath, amsthm}
\usepackage{amssymb}
\usepackage{amsfonts}
\usepackage{times}
\usepackage{latexsym}
\usepackage{amsmath, amsthm}
\usepackage{amssymb}
\usepackage{amsfonts}

\usepackage{tabularx}

\usepackage{tabularx}       
\usepackage{verbatim}       
\usepackage{fancyvrb}       
\usepackage{listings}       
\usepackage{enumitem}       
\usepackage{amsmath}        
\usepackage{geometry}       

\usepackage{booktabs}       
\usepackage{float}          
\usepackage{caption}        

\usepackage[most]{tcolorbox}
\usepackage{floatpag}

\usepackage{longtable}
\usepackage{listings}
\usepackage{booktabs, listings, enumitem}
\usepackage{listings}
\usepackage{graphicx}
\usepackage{adjustbox}

\definecolor{promptblue}{RGB}{200, 230, 255}
\definecolor{exemplargreen}{RGB}{220, 255, 220}
\definecolor{protocolyellow}{RGB}{255, 255, 200}

\setlist{nolistsep}

\usepackage{booktabs,  tcolorbox, listings, etoolbox, tabularx}
\tcbuselibrary{listings, skins, breakable}

\usepackage{fontawesome5}

\usepackage{amsmath,amsfonts,bm}

\def\eqref#1{equation~\ref{#1}}

\def\1{\bm{1}}

\DeclareMathAlphabet{\mathsfit}{\encodingdefault}{\sfdefault}{m}{sl}
\SetMathAlphabet{\mathsfit}{bold}{\encodingdefault}{\sfdefault}{bx}{n}

\definecolor{taskblue}{RGB}{0,99,177}
\definecolor{refgreen}{RGB}{0,150,85}
\definecolor{subviolet}{RGB}{131,76,190}
\definecolor{templateblue}{RGB}{1, 128, 134}
\definecolor{refgreenDark}{RGB}{0, 90, 45}
\definecolor{analysisblue}{HTML}{1E90FF} 
\definecolor{algoyellow}{HTML}{A0522D}   

\lstdefinestyle{codestyle}{
  language=Python,
  basicstyle=\footnotesize\ttfamily,
  frame=single,
  numbers=left,
  numberstyle=\tiny,
  xleftmargin=1.5em,
  framexleftmargin=1.5em,
  keywordstyle=\color{taskblue},
  commentstyle=\itshape\color{gray},
  stringstyle=\color{orange},
  showstringspaces=false,
  breaklines=true,
  tabsize=2
}
\definecolor{kwblue}{HTML}{005CFF}       
\definecolor{strred}{HTML}{B80034}    
\definecolor{codebg}{RGB}{245,248,250}
\definecolor{argsc}{RGB}{0,128,128}
\definecolor{codegreen}{HTML}{189399}
\definecolor{codebluedark}{HTML}{163D77}
\definecolor{codebluemedium}{HTML}{2F67B2}
\definecolor{codebluelight}{HTML}{7D9DCC}
\definecolor{codebluebg}{HTML}{F4F8FF}

\usepackage{listings}
\lstdefinestyle{beforeoptimcode}{
  language=Python, numbers=left, frame=none,
  numberstyle   = \tiny,        
  numbersep     = 3pt,          
  xleftmargin   = 2pt,          
  framexleftmargin = 0pt,       
  basicstyle=\ttfamily\footnotesize, keywordstyle=\color{subviolet},
  breaklines=true, columns=fullflexible
}

\lstdefinestyle{cppcontest}{
  language=C++,
  basicstyle=\ttfamily\fontsize{7}{8.4}\selectfont,
  keywordstyle=\bfseries\color{codebluedark},
  commentstyle=\itshape\color{codebluelight},
  stringstyle=\color{codebluemedium},
  numberstyle=\tiny\color{codebluelight},
  numbers=left,
  numbersep=6pt,
  showstringspaces=false,
  breaklines=true,
  columns=fullflexible,
  keepspaces=true,
  frame=single,
  framerule=0.4pt,
  rulecolor=\color{codebluelight},
  backgroundcolor=\color{codebluebg},
  xleftmargin=10pt,
  framexleftmargin=8pt,
  aboveskip=8pt,
  belowskip=8pt,
  tabsize=2,
  upquote=true
}

\usepackage[normalem]{ulem} 
\usepackage{multirow}
\usepackage{nicematrix} 
\usepackage{pgfplots}   
\usepackage{xcolor}     
\usepackage{array}      

\usepackage{arydshln}
\definecolor{code-highlight-blue}{HTML}{2696f0}
\definecolor{code-highlight-green}{HTML}{7eb547}
\definecolor{code-highlight-yellow}{HTML}{fdcc3b}
\definecolor{code-highlight-purple}{HTML}{ab4abb}
\definecolor{code-highlight-red}{HTML}{f3473a}
\definecolor{code-highlight-orange}{HTML}{fe970c}

\DeclareCaptionLabelFormat{figsub}{Figure~\thefigure(\alph{subfigure})}

\usepackage{pgfplots}
\usepackage{subcaption}

\usepackage{marvosym}

\usepackage{wrapfig}

\title{GrandCode: Achieving  Grandmaster Level in Competitive Programming via Agentic Reinforcement Learning }

\author{{\fontsize{11}{13}\selectfont Ornith Team}}

\date{}

\begin{document}

\maketitle
\begin{abstract}
Competitive programming remains one of the last few human strongholds 
in coding against AI.
The best AI system to date still underperforms the best humans competitive programming:
 the most recent best result, Google's Gemini~3 Deep Think, attained 8th place even not being evaluated under live competition conditions.

In this work, 
we introduce GrandCode, a multi-agent RL system designed for  competitive programming. 
The capability of GrandCode is attributed to two key factors:
(1) It orchestrates a variety of agentic  modules (hypothesis proposal, solver, test generator, summarization, etc) and jointly improves them through post-training and online test-time RL;
 (2) We introduce Agentic GRPO specifically designed for multi-stage agent rollouts with delayed rewards and the severe off-policy drift that is prevalent in agentic RL.

GrandCode is the first AI system that consistently beats all human participants in live contests of competitive programming: 
in the most recent three Codeforces live competitions, i.e., 
Round~1087 (Mar 21, 2026), 
Round~1088 (Mar 28, 2026), and 
Round~1089 (Mar 29, 2026),
GrandCode placed first in all of them, 
beating all human participants, including legendary grandmasters.
 GrandCode shows that AI systems have reached a point where they 
 surpass  the strongest human programmers on the most competitive coding tasks.

\begin{center}
\includegraphics[width=1\linewidth]{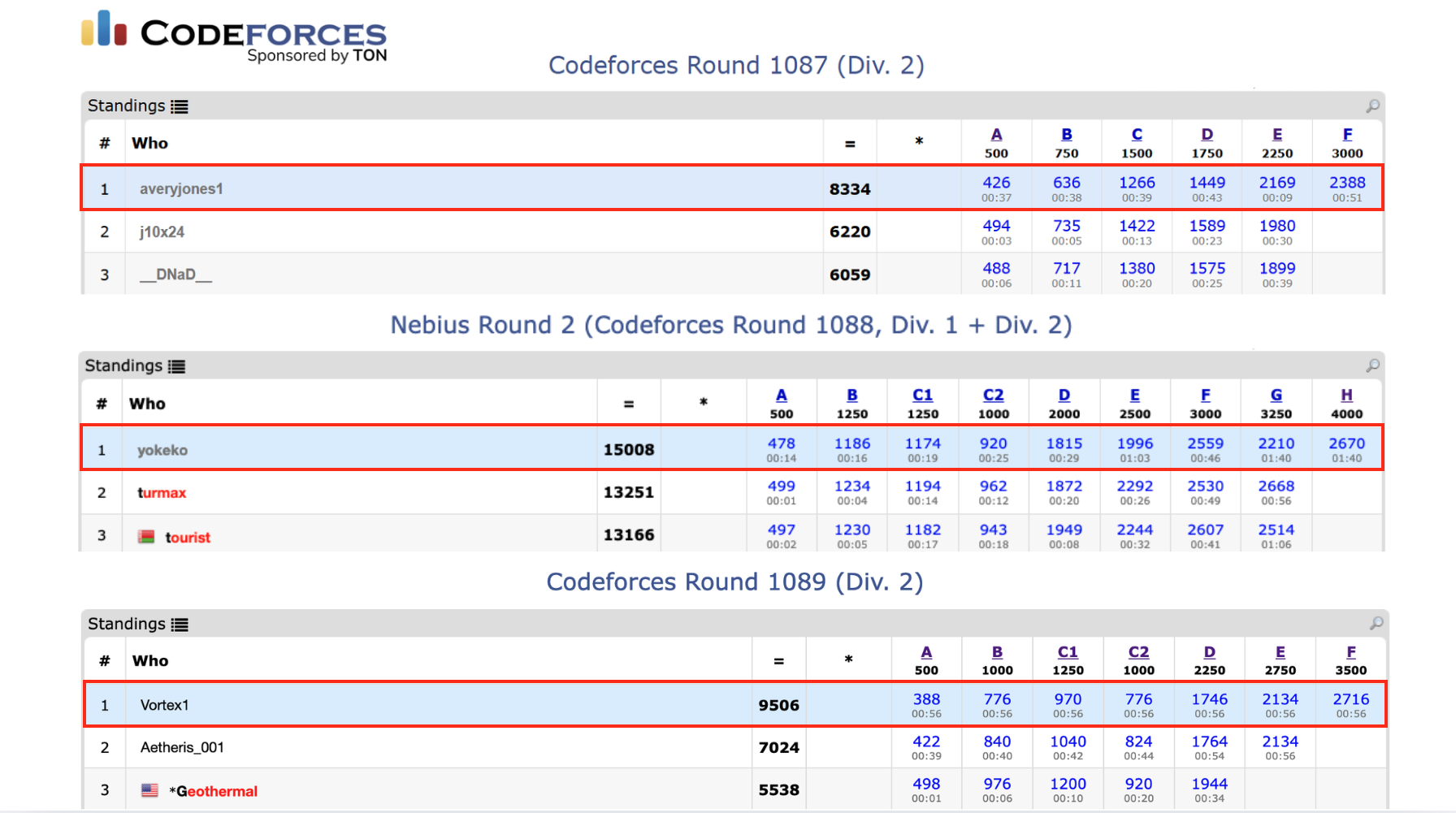}
\captionof{figure}{Codeforces standings overview for the three live
contests in which GrandCode participated. GrandCode competed as
\texttt{averyjones1} in Round~1087, \texttt{yokeko} in Round~1088, and
\texttt{Vortex1} in Round~1089, ranking first place in all three
contests and being the first to finish all tasks in each of them.}
\label{fig:cf-standings-overview}
\end{center}
\end{abstract}

\section{Introduction}

Despite rapid progress in AI for coding, the strongest current AI
systems still fall short of the best human competitors in competitive
programming. At the same time, the rapid improvement
of large language models
\citep{openai2023gpt4,openai2024gpt4o,openai2024o1,llama3,
gemini25,kimi25,qwen25,deepseekv3}
has driven substantial gains and has also spurred a growing literature
on competitive-programming benchmarks, evaluation, and datasets
\citep{codeelo2025,openai2025competitive,probench2025,
rstarcoder2025,codecontestsplus2025,hlce2025,aethercode2025,
agenticverifier2026,algoforge2025,whenelolies2026}. AlphaCode achieved a Codeforces rating of
approximately 1300, placing it in the top 54\% of competitors
\citep{li2022alphacode}; AlphaCode2 improved this to the 85th
percentile \citep{alphacode2}; and OpenAI's o3 ranks  175th globally \citep{openai2025o3}. Most
recently, Gemini~3 Deep Think attained a ranking of 8th place, though
this result was obtained on historical problems rather than under live
contest conditions.

In this work, we introduce \textbf{GrandCode}, a multi-agent reinforcement
learning system designed for competitive programming. GrandCode
orchestrates a variety of agents and  modules, and is optimized through both post-training 
and online adaptation with test-time RL in an
explicitly agentic loop: 
the hypothesis model proposes structural conjectures, 
the main solver takes the main responsibility of reasoning and solution generation, 
the summarization model maintains a compact memory of
long context, and the test-case generator produces edge test cases to challenge proposed solutions.
The goal of this design is 
to enable an agentic loop of reasoning, verification, and feedback, and
continually refining its solutions.

To address the severe off-policy issue in multi-turn agentic RL, we
introduce Agentic GRPO, a variant of Group Relative Policy Optimization
\citep{deepseekmath2024} that combines immediate
reward updates with delayed correction, enabling more effective credit
assignment under long, multi-stage rollouts and asynchronous training.

GrandCode is the first AI system to consistently surpass the best human competitors in competitive
programming under live contest conditions: 
 in the three most recent Codeforces rounds under
standard live contest conditions: Round~1087 on March~21,
Round~1088 on March~28, and Round~1089 on
March~29, 2026, GrandCode placed first in all three contests,
outperforming every human participant, including multiple top-ranked legendary grandmasters.

The rest of this paper is organized as follows.
\begin{itemize}[leftmargin=1.5em]
\item Section~\ref{sec:cf-results} describes the Codeforces evaluation
setting and GrandCode's contest results.
\item Section~\ref{sec:system-overview} presents the overall system
design.
\item Section~\ref{sec:test-case-generation} describes our test-case
generation pipeline.
\item Section~\ref{sec:hypothesis-generation} introduces the
hypothesis-generation stage.
\item Section~\ref{sec:post-training} describes continued training and
supervised fine-tuning.
\item Section~\ref{sec:multi-component-rl} presents multi-component RL
orchestration.
\item Section~\ref{sec:rl-infrastructure} describes the supporting RL
infrastructure.
\end{itemize}

\section{Codeforces Competition Results}
\label{sec:cf-results}
\subsection{Codeforces}
Codeforces\footnote{\url{https://codeforces.com/}} is one of the most prominent platforms for competitive
programming and hosts frequent public contests with large and highly
skilled participant pools. In a typical round, participants are given a
sequence of problems with increasing difficulty and must solve them as
quickly as possible under strict time and memory limits. Many rounds are
organized by division: Div.~1 typically targets higher-rated
participants, Div.~2 targets a broader pool of lower-rated participants,
and Div.~1+2 rounds combine both groups in a shared contest.
Each solution is submitted to an online judge that evaluates the code on
hidden test cases and returns only limited feedback, such as whether the
submission is wrong or exceeds the time limit. A successful submission
must therefore be both {\bf correct} and {\bf efficient}, and earlier
accepted submissions receive better scores. Problem statements are often
long, combine narrative description with precise constraints and
input-output specifications, and may also include figures, as
illustrated in Figure~\ref{tab:cf-full-problem}.

It is worth noting that
Codeforces has policies against AI-generated content, and accounts suspected of using AI face removal. 
High-ranking accounts in the contest are under even tighter scrutiny. 
To get the final score for the full version, we wait 
until the human participants have nearly finished the task before submitting the full version. 

\subsection{Participating Results}
GrandCode participated in the three most recent Codeforces live
competitions under the contestant IDs \texttt{averyjones1} in
Round~1087, \texttt{yokeko} in Round~1088, and \texttt{Vortex1} in
Round~1089.

\begin{table}[h]
\centering
\begin{tabular}{llllll}
\toprule
Round & Div. & Date  & Time (UTC+3) & Duration& ID \\
\midrule
1087 & 2 & Mar.~21, 2026  & 17:35--19:35 & 02:00:00& averyjones1 \\
1088 & 1+2 & Mar.~28, 2026  & 17:45--20:15 & 02:30:00 & yokeko\\
1089 & 2 & Mar.~29, 2026  & 17:35--19:50 & 02:15:00 & Vortex1\\
\bottomrule
\end{tabular}
\end{table}
We report two scores:
 $S(\mathrm{separate})$, which is obtained by summing the scores of
tasks at the time they are completed, i.e., by submitting each solution
as soon as it is ready. By {\it separate}, we mean that the submissions are made independently, 
with submission and standings details shown in Figures~\ref{fig:ind-scores-1087-1088} and~\ref{fig:ind-scores-1089};
and $S(\mathrm{joint})$, which is the score
based on the full set of submissions in a single account, as shown in
Figure~\ref{fig:cf-standings-overview},
with submission and standings details shown in Figure~\ref{fig:submission-details}.
In practice,
$S(\mathrm{joint})$ is strictly lower than $S(\mathrm{separate})$ because
of waiting time. 

It is worth noting that this can also lead to multiple-submission
penalties that are not reflected. We believe this effect is small
because (1) all tasks are solved within at most four submission attempts,
and (2) in the Codeforces scoring system, the penalty for multiple
attempts is small compared with the reward for early accepted
submissions.

\begin{table}[h]
\centering
\begin{tabular}{llll}
\toprule
Round  & $S(\mathrm{separate})$ & $S(\mathrm{joint})$ & Finish time \\
\midrule
1087  & 9269 & 8334 & 00:51:11 \\
1088  & 16511 & 15008 & 01:40:35 \\
1089  & 11596 & 9506 & 00:56:43 \\
\bottomrule
\end{tabular}
\end{table}

GrandCode achieved the best score in all three contests and was also the
first to finish all tasks in each of them. The corresponding
$S(\mathrm{separate})$ scores were 9269, 16511, and 11596 for
Rounds~1087, 1088, and 1089, respectively, and the corresponding
$S(\mathrm{joint})$ scores were 8334, 15008, and 9506.
Details for submissions and standings for {\it joint} is shown in Figiure \ref{fig:cf-standings-overview}.

\section{System Overview}
\label{sec:system-overview}

The proposed sytesm combines three learned policies and the test-case
generation module:
\begin{enumerate}[leftmargin=1.5em]
\item \textbf{Main solver $\pi_{\mathrm{main}}$} corresponds to the core
policy that generates reasoning traces and code.
\item \textbf{Hypothesis model $\pi_{\mathrm{hypothesis}}$} 
proposes intermediate conjectures or structural properties, 
which will be
verified
them on small instances.  Accepted hypothesis will be  injected into the prompt for Main solver. 
\item \textbf{Summarization model $\pi_{\mathrm{summary}}$} 
compresses very long reasoning traces so that hard problems remain
tractable in later RL stages.
\item \textbf{Test-case generation}
constructs adversarial tests, solution-attack tests, and large-size
stress cases to evaluate candidate programs before submission.
\end{enumerate}

The overall workflow has two phases:
\begin{enumerate}[leftmargin=1.5em]
\item \textbf{Post-training}, which can further be divided into three
substages:
\begin{enumerate}[leftmargin=1.5em]
\item {\bf Supervised fine-tuning} on high-quality (question, thinking,
solution) triples. Given (question, solution) pairs, we generate
reasoning traces for data expansion and use the resulting triples for
supervised fine-tuning, together with auxiliary components such as
$\pi_{\mathrm{hypothesis}}$ and
$\pi_{\mathrm{summary}}$.
\item {\bf Multi-component reinforcement learning} to jointly optimize the
system, enabling the main solver and auxiliary components to
collaborate more effectively under the final objective.
\end{enumerate}

\item \textbf{Test-time / live-contest solving.} During this stage, 
 the model solves the current problem instance using difficulty-aware routing, 
 direct generation for easy cases, and an online {\bf test-time RL} loop with verification feedback  for harder cases.
 \end{enumerate}

\begin{figure}[t]
\centering
\includegraphics[width=\linewidth]{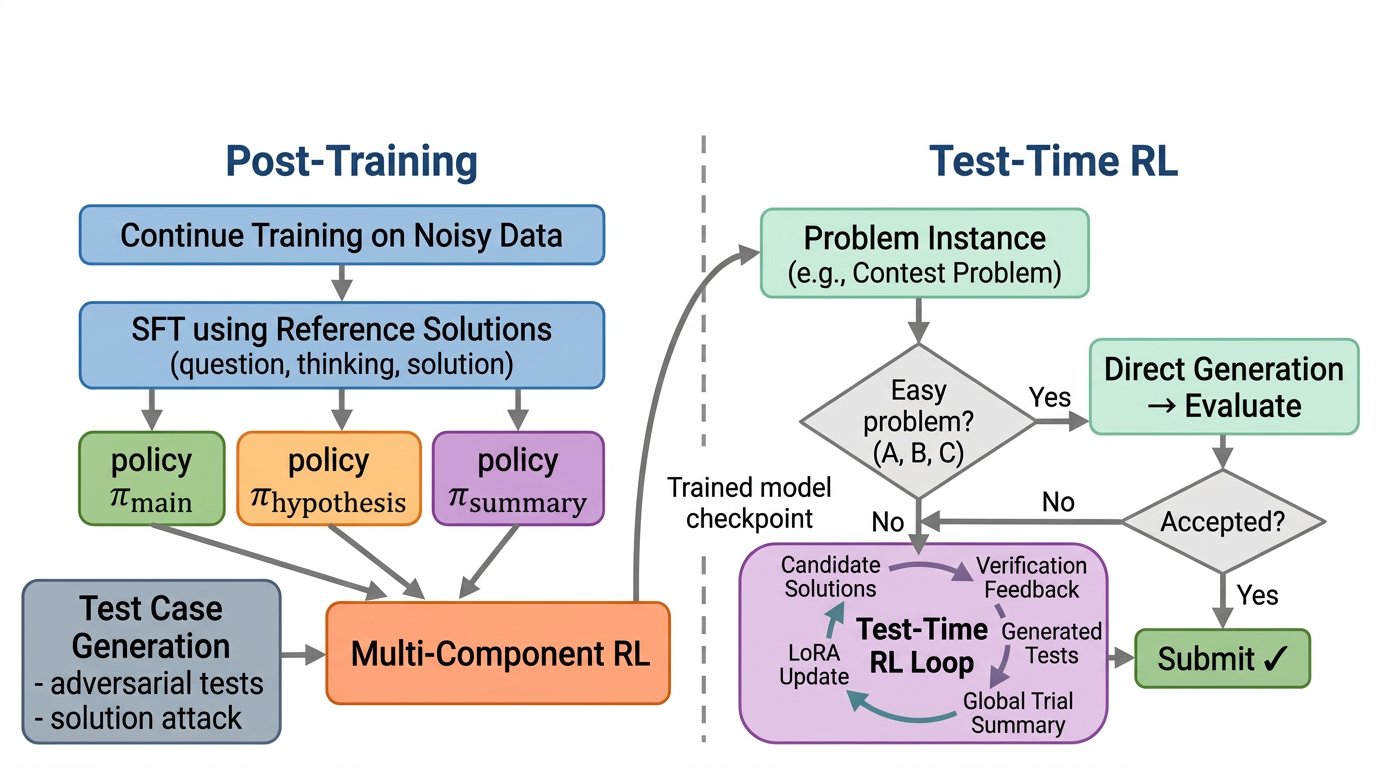}
\caption{Overview of the full pipeline. In post-training, we continue training
  on noisy competitive-programming data, perform supervised fine-tuning on
  reference solutions, train auxiliary 
  hypothesis generation policy $\pi_{\mathrm{hypothesis}}$ 
  and summarization policy $\pi_{\mathrm{summary}}$ 
  and jointly optimize the
  system with multi-component RL. At
  test/online-contest time, the model uses direct generation for easy cases, and an online
  test-time RL loop for harder
  cases.}
\label{fig:pipeline}
\end{figure}

Figure~\ref{fig:pipeline} shows the full pipeline.

\paragraph{Difficulty-based routing} We fine-tune a lightweight
classifier to assign each task to one of five difficulty levels, where
Level~1 denotes the easiest problems and Level~5 the hardest.

\section{Test Case Generation}
\label{sec:test-case-generation}
In programming contests, the real judge
test cases are hidden, so we must construct its own test cases and pass them before the submission. 
Therefore, we need to
 generate edge cases 
that can  expose logical bugs, boundary
failures, and incorrect complexity assumptions before submission.

Test case generation  faces two main challenges. First, it is
difficult to generate genuinely {\bf adversarial test cases} that
expose subtle logical errors rather than merely checking superficial
correctness. Second, {\bf large-size test cases} are hard to use for
verification, because in many settings the only trusted solver
available to us is a brute-force implementation, and that solver times
out on large inputs, which makes we don't have gold outputs to compare with for large-size inputs. 

\subsection{Adversarial Test Case Generation} We adopt two strategies for adversarial test generation:
{\bf difference-driven test generation} and {\bf solution attack}.

\paragraph{Difference-driven Test Case Generation}
If 
a test
case can expose a difference between two solutions, it is very likely an adversarial case.

Given a list of candidate solutions sampled during training, we
iteratively generate candidate test cases by prompting Claude, GPT,
DeepSeek~V3, and Kimi~2.5 by  
to produce
 inputs that are likely to reveal corner cases. No candidate
solution is provided to the LLMs at this stage. We further employ a
generator-validator framework inspired by CodeContests+
\citep{codecontestsplus2025}, in which an additional LLM validator is
used to filter or refine the generated tests.
We then run the generated test cases on all sampled solutions and check
whether any test induces different outputs between them. Test
cases that trigger such differences are preserved.
Whenever such a case is found, we feed it back to the LLM and ask it to
describe what edge condition the input may be triggering, and then
generate additional tests of a similar kind. In this way, the test pool
is gradually enriched with more informative cases, and by the end of
optimization for a problem we are able to maintain a strong
problem-specific test suite. 
It is worth noting that  this idea is akin to the spirit to the agentic
verification approach \cite{agenticverifier2026}. 

\paragraph{Solution Attack}

In the training set, where a gold solution is available, we further
generate adversarial tests by directly comparing the gold solution with
each candidate solution. We feed both solutions to an LLM and ask it to
analyze their differences, identify potential bugs in the candidate
solution, and propose edge cases that are likely to expose those bugs.
We carry out this process in a multi-turn conversation mode, allowing
the model to iteratively refine its hypotheses and search for stronger
adversarial tests.
Each generated test case is then executed on both the gold solution and
the candidate solution to verify whether it indeed induces a behavioral
difference. Test cases that successfully separate the two solutions are
preserved, since they provide direct evidence of a
real bug in the candidate solution.

\subsection{Test-time Strategies} When we handle a specific contest task, we use the similar
online strategy: 
We  prompt the fine-tuned model to generate multiple
adversarial test cases conditioned on each generated solution, and
after every few candidate solutions, we regenerate
part of the test set and refresh the evaluation pool. Test examples
that trigger a difference are kept, since they are highly
informative indicators of edge cases.
 A
solution is submitted only if it passes all of these  test
cases.

\subsection{Results on Real Codeforces Problems}
We evaluate 
test case genearation
 on 50 real Codeforces problems by submitting
solutions to the real Codeforces website and using the real system as the final criterion. We check whether a solution that passes our
generated test suite also passes the hidden official tests.
Table~\ref{tab:real-cf-tests} summarizes the results. The pass
count increases from 42 to 48 after applying difference-driven test
case generation and solution attack. For the remaining two failures, we
further incorporate submission feedback and continue generating
additional test cases online, which raises the pass count to all of 50 tests.

\begin{table}[t]
\centering
\begin{tabular}{lcc}
\toprule
Stage & Passed & Total \\
\midrule
Base test suite & 42 & 50 \\
After difference-driven generation + solution attack & 48 & 50 \\
After submission feedback + continued online generation & 50 & 50 \\
\bottomrule
\end{tabular}
\caption{Results on 50 real Codeforces problems using the Codeforces
judge as the final criterion.}
\label{tab:real-cf-tests}
\end{table}

\subsection{Large-size Test Cases}

In real-time contest, feedback from the submission system can also be harnessed 
to generate test cases with larger input sizes.
A failure due to time limit exceeded often suggests
that the generated code snippet may be logically correct but computationally
inefficient. If such the code snippet is faster than the
brute-force baseline, we can use it as a solver on test
cases with larger input sizes, allowing evaluation on inputs that are
too large for brute force.
\begin{figure}[t]
\centering
\includegraphics[width=1.1\linewidth]{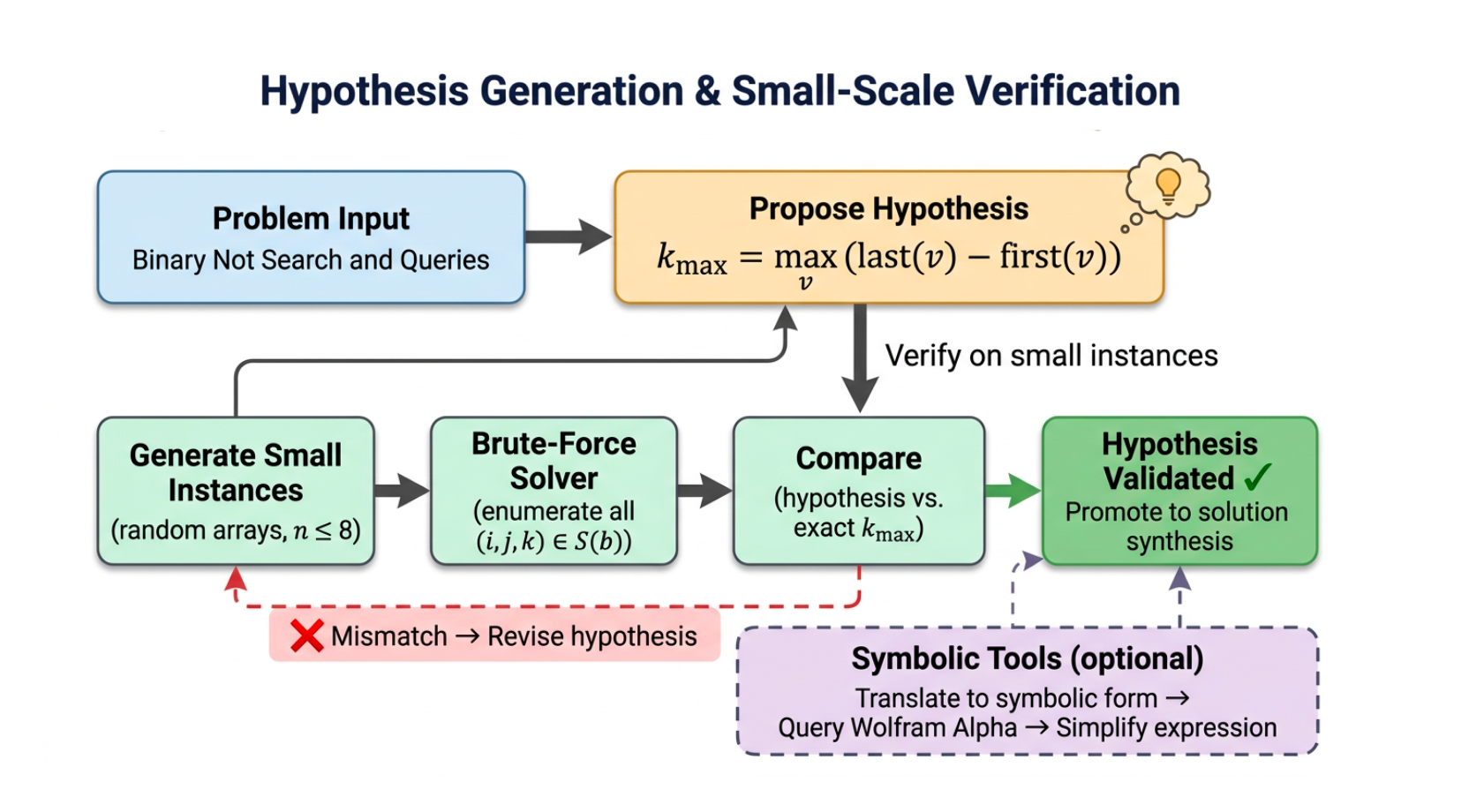}
\caption{Hypothesis generation and small-scale verification.
  The agent first proposes a compact characterization
  ($k_{\max} = \max_v (\mathrm{last}(v) - \mathrm{first}(v))$), then
  generates small random instances, computes the exact
  answer via a brute-force solver that enumerates all tuples
  $(i,j,k)\in S(b)$, and compares against the hypothesized value.
  A mismatch triggers hypothesis revision and only validated hypotheses
  are promoted to solution synthesis.}
\label{fig:hypo-gen}
\end{figure}

\section{Prelude: Hypothesis Generation}
\label{sec:hypothesis-generation}
\subsection{Overview}
The first stage of the agent workflow is {\bf hypothesis generation} and
{\bf small-scale verification}. Before attempting full solution synthesis,
we need to propose intermediate claims, structural properties,
or compact mathematical characterizations of the problem, and then
checks whether they are valid on small instances. For example, the agent may verify that
whether the problem is a dynamic-programming problem in nature, or confirm that the input graph satisfies a structural
property such as being undirected. These checks can be easily performed on
small-scale inputs using the brute-force algorithm.

This stage can also invoke symbolic tools. When a problem 
or a broken-down subproblem can 
can be fully translated 
to
 a symbolic form, we will query an external engine
such as Wolfram Alpha\footnote{\url{https://www.wolframalpha.com/}} to simplify or solve the resulting expression.
This is especially useful when the main difficulty lies in discovering
the right formula or invariant rather than in implementation details.

As a concrete example, consider the Codeforces problem ``Binary Not
Search and Queries.'' We may first hypothesize that
$k_{\max} = \max_v (\mathrm{last}(v) - \mathrm{first}(v))$. 
Next we ask the model to
 generate many small
inputs, computes the exact answer with a brute-force solver, and
checks whether the conjectured formula matches the true value on those
instances. When a mismatch is found, the counterexample is fed back to
the model together with the failing hypothesis, and the model is asked
to propose a revised conjecture; this verify-and-revise loop repeats
until the hypothesis is consistent with all small-scale tests. 
This process is illustrated in Figure~\ref{fig:hypo-gen}.

Hypotheses that survive this iterative validation are injected into the
prompt of the main solving thread, providing verified structural
insights that guide subsequent solution synthesis. 
The tools used in this stage include shell execution, Wolfram Alpha,
and web search. For search in particular, we apply a filter to avoid
directly retrieving historical answers, editorials, or hints for the
target problem.
\subsection{Training Hypothesis Generation Model}
We use a lightweight model for this component rather
than allocating a large model to every instance. We denote this model
by $\pi_{\mathrm{hypothesis}}$.

\paragraph{SFT} We first train  $\pi_{\mathrm{hypothesis}}$  with supervised fine-tuning. To construct SFT
data, we use Claude to generate multiple candidate hypotheses given
only the problem statement, and retain those that are verified to be
correct. For problems that come with editorials or reference solutions,
we additionally provide these materials to Claude and ask it to produce
hypotheses implied by the official reasoning, again retaining only the
correct ones. The resulting verified hypotheses are then used as
supervision for SFT.

\paragraph{RL} After SFT, we further optimize $\pi_{\mathrm{hypothesis}}$ with
reinforcement learning using GRPO.
At the current stage, the reward for a generated hypothesis $h$ is the
proportion of verification tests it passes:
\begin{equation}
r_{\mathrm{verify}}(h)=\frac{1}{|T|}\sum_{t \in T}\mathbf{1}[h\text{ is
correct on }t].
\end{equation}
At the current stage, $\pi_{\mathrm{hypothesis}}$ is trained
independently for simplicity. However, this is inherently imperfect,
since generating a correct hypothesis does not necessarily imply that it is helpful to generate the correct solution. In the full system,
$\pi_{\mathrm{hypothesis}}$ will later be jointly trained with the
overall solver in the RL training stage for the latter.

\subsection{Clue Finding Using OEIS}

We also employ a side route that does not
involve any additional model training. We first 
compute outputs for a few small values of $N$, and query the On-Line
Encyclopedia of Integer Sequences (OEIS, \url{https://oeis.org/}) with
the resulting sequence to search for useful clues. If the lookup is
successful, the returned pattern, formula, or related structural hint
is included in the prompt for subsequent solving.

\section{Post Training with Continue Training and SFT}
\label{sec:post-training}
\paragraph{Benchmarking Existing Models}
We select 100 questions, equally distributed across the five
difficulty categories, and evaluate proprietary 
models on this benchmark. Table~\ref{tab:benchmark-existing-models}
reports three complementary metrics for each model: the overall accept
rate, the number of hardest Level~5 problems solved, and a
difficulty-weighted score that assigns weights $1,2,3,4,5$ to Levels
1--5, respectively. Overall, current frontier models achieve an overall accept
rate of roughly 70\%--75\% and solve about 35\%--40\% of Level~5
problems.

\begin{table}[t]
\centering
\begin{tabular}{lccc}
\toprule
Model & Accept Rate & Level 5 Solved & Weighted Score (0-100) \\
\midrule
Gemini 3.1 Pro & 75\% & 7/20 & 64.3 \\ 
Claude Opus 4.6 & 73\% & 8/20 & 63.7 \\
GPT-5.4 & 72\% & 7/20 & 63.0 \\
Kimi K2.5 & 65\% & 5/20 & 53.3 \\
DeepSeek V3.2 & 65\% & 4/20 & 52.7 \\
Qwen 3.5-397B & 64\% & 4/20 & 52.3 \\
\bottomrule
\end{tabular}
\caption{Accept rates, Level 5 correct answers (out of 20), and scaled weighted scores on 100 benchmark questions. Questions are equally distributed across five difficulty categories. Weighted scores apply a 1-5 multiplier based on difficulty, normalized to a 0-100 scale.}
\label{tab:benchmark-existing-models}
\end{table}

\subsection{Continue Training on Noisy Data}
For continued pretraining, we first
collect a broad seed set of competitive-programming problems from TACO
\citep{taco2023}, LeetCode \citep{leetcode}, USACO \citep{usaco},
CodeContests \citep{li2022alphacode}, IOI \citep{ioi}, and additional
problems crawled from various online sources.  
we randomly convert 20\% of the continued-pretraining examples into
hypothesis-conditioned cases, in which a hypothesis
generated by $\pi_{\mathrm{hypothesis}}$
 is 
 incorporated in the prompt
  before we generate the detailed thinking process.
At this stage, the data can be noisy because it is partly generated,
and some synthesized reasoning traces or answers may be incorrect.
However, the goal of continued pretraining is primarily to improve the
model's general competitive-programming ability rather than to provide
precise supervision. We leave more fine-grained filtering and
high-quality supervision to the later SFT stage.

\subsection{SFT using Reference Solutions}
For SFT, we focus only
on non-synthetic problems with reference solutions, some of which also come with hints. Our goal is to generate a high-quality
thinking process $c$ that can plausibly lead to the provided reference
solution.
With such tuples, we can directly perform SFT by training the model $G_{\mathrm{main}}$ to
first predict $c$ given $x$, and then predict $y$ given $(x, c)$, denoted by $p(y, c \mid x)$. 

\paragraph{Solution Matching} For each question $x$, we prompt
GLM~4.5,
Kimi~K2.5, and DeepSeek~V3.2 multiple times to generate both a reasoning trace $c$ and a
solution $y$. We then compare the generated solution against the gold
solution. If the generated solution is equivalent to the gold solution,
we keep the associated reasoning trace. Even when the implementation is
not textually identical, we still retain it if it is comparable to or better than the reference solution in terms of efficiency, since many problems admit multiple valid solutions. 

It is worth noting that
this strategy is effective for relatively easy tasks, where we can often recover a solution close to the gold one. For harder
tasks, however, the generated solution frequently fails to match the
gold solution, making this simple filtering procedure insufficient.

\paragraph{Finding Optimal Thinking Trace $c$ for Hard Problems}
We denote the full forward model by $\pi_{\mathrm{main}}$, and its
original checkpoint by $\pi_{\mathrm{main}}^{(0)}$.
We then select $c_i$ based on the following score:
\begin{equation}
s(c_i) =
\alpha \cdot \frac{\log p_{\pi_{\mathrm{main}}}(c_i, y \mid x)}{|c_i| + |y|}
+
\beta \cdot \frac{\log p_{\pi_{\mathrm{main}}^{(0)}}(x \mid c_i)}{|x|},
\end{equation}
Intuitively, the
first term favors reasoning contexts that make the target reasoning and
solution likely under the post-trained model, while the second term
favors contexts that remain predictive of the original question. The second
term is essential for distinguishing genuinely useful reasoning traces
from degenerate ones that merely copy or reveal the answer. 
At the end of this procedure, the tuples are combined into the
the SFT training data.

\subsection{Training Summarization Model}

For hard questions, the thinking trace can easily exceed 100K tokens,
which makes both inference and the later RL stage extremely computationally
expensive. Additionally, extremely long traces are also harder to optimize with RL. We therefore first
train a separate summarization model $\pi_{\mathrm{summary}}$.

Given a long reasoning trace $c$, we partition it into chunks
$c^{(1)}, c^{(2)}, \ldots, c^{(n)}$, so that the full example is
represented as $(x, c^{(1)}, c^{(2)}, \ldots, c^{(n)}, y)$. The
summarization model maintains a progressive summary state
$s_1, s_2, \ldots, s_n$, where
$\pi_{\mathrm{summary}}(c^{(1)}) \rightarrow s_1$,
$\pi_{\mathrm{summary}}(s_1, c^{(2)}) \rightarrow s_2$,
and in general
$\pi_{\mathrm{summary}}((s_{t-1}, c^{(t)}) \rightarrow s_t$. Intuitively, $s_t$ is a compact
summary seen up to chunk $t$. 

There has been existing work that integrates summary generation into
training mostly in an end-to-end fashion, including MemAgent
\citep{memagent2025}, Agentic Memory \citep{agenticmemory2026}, and
Composer~2 \citep{composer2}. In these approaches, summarization 
usually shares the parameters with the main model, and is optimized jointly with the main model in an end-to-end fashion
based on the final answer. 

While this
end-to-end formulation is natural, its reward is relatively sparse,
since supervision is dominated by the terminal outcome of the full
trajectory. By contrast, we first take advantage of the existing SFT
data to learn as strong a summarization policy as possible, rather than
directly entering the RL stage and relying only on the final reward as
the training signal. We train the summarizer progressively in
multiple stages. This gives
denser intermediate supervision in the early stage of training.

\paragraph{Stage 1}
We first train each local summarization step with RL.  Concretely, for a sampled summary $s_1$, we define the score
as:
\begin{equation}
\mathrm{score}(s_1) =
\alpha \, \frac{\log p\!\left(c^{(2)}, \ldots, c^{(n)}, y \mid s_1, x\right)}
{\left|c^{(2)}, \ldots, c^{(n)}\right| + |y|}
+
\beta \, \frac{\log p\!\left(c^{(1)} \mid s_1\right)}
{\left|c^{(1)}\right|}
\end{equation}

The first term encourages $s_1$ to preserve information needed for the
later chunks and the final answer, while the second term penalizes
reconstruction error by encouraging $s_1$ to retain enough information
to reconstruct the original chunk.

We optimize this policy with GRPO. Given a group of rollout rewards
$\{r_i\}_{i=1}^G$, we normalize them within the group as
$A_i = (r_i - \mathrm{mean}(r))/\mathrm{std}(r)$ and update the policy
using these relative advantages.
We use analogous objectives for later transitions
$(s_{t-1}, c^{(t)}) \rightarrow s_t$. This stage teaches each summary
state to preserve the information necessary for continuing the long
reasoning process. 

\paragraph{Stage 2}
We then train the full progressive chain end to end. We sample the
entire summarization trajectory
$c^{(1)} \rightarrow s_1$,
$(s_1, c^{(2)}) \rightarrow s_2$,
$\ldots$,
$(s_{n-1}, c^{(n)}) \rightarrow s_n$,
and use the final answer likelihood, e.g. the normalized $\log p(y \mid s_n)$, as the
terminal reward for GRPO.

\paragraph{Stage 3} 

Finally, $\pi_{\mathrm{summary}}$ is integrated into the full RL training
of $\pi_{\mathrm{main}}$, so that summarization and downstream solving can
be optimized jointly in the overall pipeline, as will be described in
detail in Section~\ref{RL}. 

In addition, data with summaries
is mixed into the SFT training data, so that $\pi_{\mathrm{summary}}$
becomes familiar with both settings where summaries are present and
settings where they are not.

\section{Multi-Component RL Orchestration}
\label{sec:multi-component-rl}
\label{RL}
The final stage involves multiple components for RL training, including
$\pi_{\mathrm{main}}$, $\pi_{\mathrm{hypothesis}}$, and
$\pi_{\mathrm{summary}}$. 
We choose
multi-component RL for three main reasons: (1) \textbf{Smaller auxiliary
models:} $\pi_{\mathrm{hypothesis}}$ and $\pi_{\mathrm{summary}}$ use
smaller models, which saves substantial compute resources;
(2) \textbf{Stronger initialization:} $\pi_{\mathrm{hypothesis}}$ and $\pi_{\mathrm{summary}}$  are separately
pretrained in earlier stages and are already reasonably strong before
the final joint RL stage, so they require only moderate joint tuning
together with $\pi_{\mathrm{main}}$; (3) \textbf{Better disentanglement:}
separate training makes it easier to isolate the influence of different
modules: if one uses only a single final solution reward 
for $\pi_{\mathrm{summary}}$ and $\pi_{\mathrm{main}}$
and the
result is poor, it is unclear whether the failure comes from a bad
summary or a bad solver. This modular design therefore makes the final
RL stage substantially more compute-efficient and easier to optimize.

\subsection{Reward for $\pi_{\mathrm{main}}$}
Reward evaluation of a generated code snippet consists of
three stages:

\begin{enumerate}[leftmargin=1.5em]
\item \textbf{Executability.} If a generated code snippet cannot be compiled
or executed, it receives score $0$.
\item \textbf{Correctness.} Correctness is measured by comparing
the code snippet output with a reference output. When a gold solution is
available, as in training-time evaluation, we use the gold solution to
produce the reference output. 
At test time, when no gold solution is available, we compare against
the output of a brute-force solver on small inputs. If a code snippet
fails these correctness checks, it receives score $0$.
\item \textbf{Efficiency.} For a code snippet that pass correctness checking,
we compare runtime against a brute-force baseline on the generated
tests, whose scale doesn't has to be small. For each test case, the reward is defined as the speedup over
the brute-force algorithm; if the generated code times out, we
assign a score of $0.1$ for that test case. The final score is the
average over all test cases.
\end{enumerate}

A detailed formalization in math in shown in Appendix \ref{CodeReward}.

\paragraph{Length Penalty based on Difficulty}
More difficult questions typically require longer chains of reasoning,
so the thinking-length penalty should depend on the  problem
difficulty. Recall that we fine-tune a lightweight classifier that
assigns each task a difficulty level $d \in \{1,2,3,4,5\}$, where
larger $d$ indicates a harder problem. We therefore allocate a larger
thinking-token budget to harder tasks. Let $l$ denote the thinking
length, $B$ the base budget for Level~1 problems, and $\alpha$ the
budget growth factor per difficulty level. The difficulty-dependent
budget is $B \cdot \alpha^{(d-1)}$, and the penalty is zero when the
thinking length stays within this budget and increases only when it is
exceeded:
\begin{equation}
Penalty(l,d) = \max\!\left(0,\;\frac{l - B \cdot \alpha^{(d-1)}}{B \cdot \alpha^{(d-1)}}\right).
\end{equation}
In this way, easier problems are encouraged to use short, efficient
reasoning traces, while harder problems are allowed more extensive
thinking before incurring a penalty.

\subsection{Orchestrating of $\pi_{\mathrm{main}}$ and $\pi_{\mathrm{hypothesis}}$}

We first sample multiple candidate hypotheses from
$\pi_{\mathrm{hypothesis}}$. Only hypotheses that pass all verification
tests are allowed to continue to the next stage. The reward for the
hypothesis-generation stage is
\begin{equation}
r_1(h)=r_{\mathrm{verify}}(h)=\frac{1}{|T|}\sum_{t \in T}\mathbf{1}[h\text{ is
correct on }t].
\end{equation}

For each passed hypothesis $h$, we then invoke $\pi_{\mathrm{main}}$ to
generate reasoning traces and candidate solutions conditioned on $(x,
h)$. In parallel, we also generate rollouts from $\pi_{\mathrm{main}}$
without any hypothesis, which we denote by the empty condition
$\varnothing$. Let $S(x,h)$ denote the average score of solutions
generated with hypothesis $h$ incorporated, and let $S(x,\varnothing)$
denote the average score of solutions generated without any hypothesis.

We define the total reward for  $\pi_{\mathrm{hypothesis}}$ by
\begin{equation}
r(h)=
\begin{cases}
\alpha r_1(h)+\beta \bigl(S(x,h)-S(x,\varnothing)\bigr),
& \text{if } h \text{ passes all tests},\\
\alpha r_1(h), & \text{otherwise.}
\end{cases}
\end{equation}
This design encourages $\pi_{\mathrm{hypothesis}}$ not only to generate
locally valid hypotheses, but also to produce hypotheses that
measurably improve downstream solution quality when used by
$\pi_{\mathrm{main}}$.
Importantly, $\pi_{\mathrm{hypothesis}}$ is allowed to generate
hypotheses over multiple rounds with self-correction. This iterative loop lets the hypothesis generator
recover from initial mistakes and converge toward hypotheses that
are both locally correct and globally useful.
\subsection{Orchestrating of $\pi_{\mathrm{summary}}$ and $\pi_{\mathrm{hypothesis}}$}

$\pi_{\mathrm{summary}}$ is triggered only for long reasoning
sequences. During training, we gradually increase its triggering
frequency, together with the curriculum that shifts
toward harder problems over time. 
As a result, early training focuses primarily on
$\pi_{\mathrm{hypothesis}}$ and $\pi_{\mathrm{main}}$, while later
training increasingly exposes the system to summary-dependent rollouts.
When $\pi_{\mathrm{summary}}$ is triggered, we use the same final reward
as for $\pi_{\mathrm{main}}$. Concretely, if the summary-conditioned
rollout ultimately produces code snippet $P$, then the reward assigned
to $\pi_{\mathrm{summary}}$ is the final rollout score $r(P)$ defined
for the main solver.

At the systems level, we place the large MoE solver policy
$\pi_{\mathrm{main}}$ on a dedicated distributed GPU mesh, since it is the only component that
requires both expert parallelism and long-context context parallelism.
The auxiliary policies $\pi_{\mathrm{hypothesis}}$ and
$\pi_{\mathrm{summary}}$, both implemented as smaller dense models, are
served asynchronously on separate small GPU pools. This avoids
fragmenting the main MoE mesh, keeps the main rollout/training pipeline
saturated, and allows hypothesis and summary requests to be batched
independently. Code execution, brute-force checking, and test
generation are handled by a separate CPU sandbox pool.

\begin{figure}[t]
\centering
\includegraphics[width=0.95\linewidth]{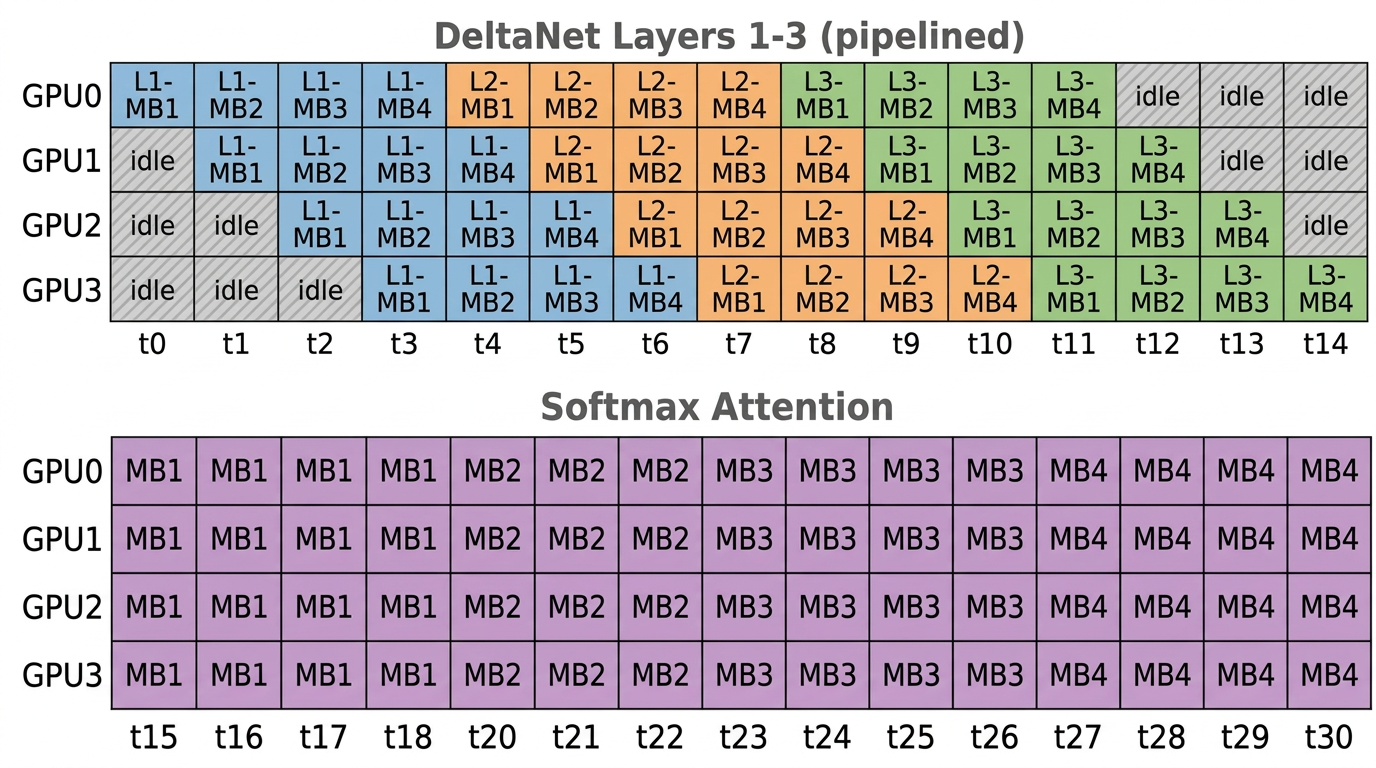}
\caption{An illustration of pipelined context parallelism for one block
  with 3~DeltaNet layers (L1, L2, L3) + 1~softmax attention layer with 4~CP ranks and
  4~micro-batches for illustration purposes.  In the DeltaNet phase, each GPU processes micro-batches (MB)
  in a staggered pipeline, passing the recurrent state forward; startup
  and drain bubbles (gray) are confined to the triangular corners.
  The softmax attention layer is executed with synchronized
  all-to-all communication at full utilization.}
\label{fig:pipelined-cp}
\end{figure}

\section{RL Infrastructure}
\label{sec:rl-infrastructure}

\subsection{Asynchronous Training}

To control the effect of earlier generated off-policy tokens, we apply a
staleness weight $w(d_t)$ that downweights tokens according to their age
$d_t$ and drops them entirely once a threshold is exceeded:
\begin{equation}
w(d_t)=
\begin{cases}
\!1, & \text{if } d_t \le K_1,\\
\!\exp\!\bigl(-\lambda(d_t-K_1)\bigr), & \text{if } K_1 < d_t \le K_2,\\
\!0, & \text{if } d_t > K_2.
\end{cases}
\label{drop}
\end{equation}
where $K_1$ is the threshold beyond which staleness is penalized and
$K_2$ is the hard max-lag threshold. We then weight the token-level
GRPO loss as
\begin{equation}
L_t=\min\!\bigl(r_t A_t,\; \mathrm{clip}(r_t,1-\epsilon^{-},1+\epsilon^{+})A_t\bigr)\cdot w(d_t),
\end{equation}
where
\begin{equation}
r_t=
\frac{\pi_{\theta}(y_t \mid x, y_{<t})}
{\pi_{\theta_t^{\mathrm{beh}}}(y_t \mid x, y_{<t})}
\end{equation}
is the token-level policy ratio, where $\theta_t^{\mathrm{beh}}$
denotes the behavior-policy version that generated token $y_t$, whose
age is $d_t$.
It is worth noting that the token-dropping rule in Eq.~\ref{drop}
works naturally together with Agentic-GRPO. Intermediate rewards might be not affected, because they are used immediately when they become
available. By contrast, delayed correction terms may be dropped if they
arrive too late and exceed the staleness threshold.

\subsection{Pipelined CP for DeltaNet+Softmax Attention Mixture}
Because thinking tends to be super long for hard problems, we primarily use context parallelism (CP) to address the long-context issue. 

Hybrid architectures that combine linear attention with softmax
attention are becoming increasingly common, balancing the efficiency of
linear attention with the stronger modeling capacity of softmax attention.
Since DeltaNet is a linear recurrent operator,
standard context parallelism is not a natural fit: CP rank $t$ cannot
start processing until CP rank $t-1$ has finished and passed along the
required recurrent state.  Inspired by pipeline
parallelism, we therefore use a pipelined CP design by
pipelining multiple batches across CP
ranks. Operationally, one block consists of
three phases: Phase~1 pipelines the three DeltaNet layers with full
overlap across micro-batches, Phase~2 inserts a synchronization barrier
before attention, and Phase~3 executes the softmax attention layer. This hybrid design preserves the
sequential semantics of DeltaNet while amortizing pipeline bubbles over
many micro-batches. In a representative setting with 4 CP ranks and 4
micro-batches, one block achieves about 90\% overall utilization, and
the efficiency further improves when RL training supplies many rollout
micro-batches.
Figure~\ref{fig:pipelined-cp} illustrates the schedule for 4 CP ranks
and 4 micro-batches.

\subsection{Others}
\paragraph{Balanced Difficulty/Length in Batching}
Thinking length is highly correlated with problem difficulty. As a
result, mixing easy and hard problem within the same batch
can lead to substantial imbalance in sequence length. This is also a
problem for data parallelism, since different DP workers may otherwise
receive batches with very different compute costs. Therefore, in both
RL training and test-time execution, we organize not only the instances
within each batch to have similar difficulty levels, but also align the
difficulty of batches across DP workers. At the beginning, when no
runtime statistics are yet available, we batch primarily by difficulty.
We then record the max completion time of its rollout, and when the
same instance is revisited in later rounds, we batch it according to the
measured completion time from the previous round. 

\paragraph{Dynamic CP} 
Because thinking length varies substantially across difficulty levels,
the optimal amount of context parallelism also differs across batches.
Using a single fixed CP size for all batches would therefore be
inefficient: easy batches may over-parallelize short sequences, while
hard batches may under-parallelize very long ones. We therefore adopt a
dynamic CP strategy  \citep{dcp2025}, adjusting the CP size based on the difficulty level for each batch.

\paragraph{Expert Routing Stability and Load Balancing}
To avoid routing instability during RL training, we freeze the router
entirely and update only the expert feed-forward parameters. This keeps
expert assignments fixed throughout RL, preventing routing drift between
rollout and training and avoiding additional instability from changing
expert load patterns.

\begin{table}[t]
\centering
\begin{tabular}{lccc}
\toprule
Micro-batches & DeltaNet Efficiency & Overall Efficiency & Bubble Overhead \\
\midrule
2 & 67\% & 87\% & 13\% \\
4 & 80\% & 90.3\% & 9.7\% \\
8 & 89\% & 94\% & $\sim$6\% \\
16 & 94\% & 97\% & $\sim$3\% \\
$N$ (large) & $N/(N+3)$ & $\to 100\%$ & $\to 0$ \\
\bottomrule
\end{tabular}
\caption{Efficiency of pipelined context parallelism for the 3DeltaNet+Softmax Attention mixture as the number of micro-batches increases.}
\label{tab:pipelined-cp-efficiency}
\end{table}

\subsection{Results}

Table~\ref{tab:test-time-rl-results} summarizes the progression from the
post-training model to full RL training and then to test-time RL.
Starting from the post-training result of 72\% accept rate, 7/20 solved
Level~5 problems, and a weighted score of 61.7, full RL training
substantially improves performance to 81\%, 13/20, and 72.3,
respectively. Applying test-time RL yields a further gain to 85\%
overall accept rate, 15/20 solved Level~5 problems, and a weighted score
of 73.5. These results suggest that offline RL training already provides
a large improvement in core problem-solving ability, while test-time RL
is especially helpful on the hardest problems.

\section{Conclusion}
In this work, we introduced GrandCode, a multi-agent reinforcement
learning system for agentic competitive programming. Its
performance comes from two main ingredients: a coordinated agentic loop
of reasoning, hypothesis generation, summarization, test-case
generation, and verification; and Agentic GRPO, which addresses delayed
rewards and severe off-policy drift in long, multi-stage agent rollouts.

Across offline benchmarks, continued training, supervised fine-tuning,
full RL, and test-time RL all contribute meaningful gains, especially on
hard problems. Under standard live contest conditions, GrandCode ranked
first in all three recent Codeforces rounds in which it participated,
outperforming all human contestants, including top legendary
grandmasters. Taken together, these results suggest that agentic
reinforcement learning, when combined with strong verification and
online adaptation, can push coding systems beyond top
human performance in real-time environments.

\bibliography{custom}

\begin{thebibliography}{41}
\providecommand{\natexlab}[1]{#1}
\providecommand{\url}[1]{\texttt{#1}}
\expandafter\ifx\csname urlstyle\endcsname\relax
  \providecommand{\doi}[1]{doi: #1}\else
  \providecommand{\doi}{doi: \begingroup \urlstyle{rm}\Url}\fi

\bibitem[{AlphaCode Team, Google DeepMind}(2023)]{alphacode2}
{AlphaCode Team, Google DeepMind}.
\newblock {AlphaCode} 2 technical report.
\newblock Technical report, Google DeepMind, 2023.
\newblock URL
  \url{https://storage.googleapis.com/deepmind-media/AlphaCode2/AlphaCode2_Tech_Report.pdf}.

\bibitem[Anonymous(2025)]{ttrl2025}
Anonymous.
\newblock {TTRL}: Test-time reinforcement learning.
\newblock \emph{arXiv preprint arXiv:2504.16084}, 2025.

\bibitem[Chan et~al.(2026)Chan, Shalaby, Wettig, et~al.]{composer2}
Aaron Chan, Ahmed Shalaby, Alexander Wettig, et~al.
\newblock Composer 2 technical report.
\newblock \emph{arXiv preprint arXiv:2603.24477}, 2026.

\bibitem[{DeepSeek-AI}(2025)]{deepseekv3}
{DeepSeek-AI}.
\newblock {DeepSeek-V3} technical report.
\newblock \emph{arXiv preprint arXiv:2412.19437}, 2025.

\bibitem[Dou et~al.(2025)Dou, Zhao, and Biswas]{algoforge2025}
Zhihao Dou, Qinjian Zhao, and Sumon Biswas.
\newblock Algoforge: Specializing code generation agents through collaborative
  reinforcement learning.
\newblock OpenReview, 2025.
\newblock ICLR 2026 withdrawn submission.

\bibitem[El-Kishky et~al.(2025)El-Kishky, Wei, Saraiva, Minaiev, Selsam, Dohan,
  Song, Lightman, Clavera, Pachocki, Tworek, Kuhn, Kaiser, Chen, Schwarzer,
  Rohaninejad, McAleese, contributors, M{\"u}rk, Garg, Shu, Sidor, Kosaraju,
  and Zhou]{openai2025competitive}
Ahmed El-Kishky, Alexander Wei, Andre Saraiva, Borys Minaiev, Daniel Selsam,
  David Dohan, Francis Song, Hunter Lightman, Ignasi Clavera, Jakub Pachocki,
  Jerry Tworek, Lorenz Kuhn, Lukasz Kaiser, Mark Chen, Max Schwarzer, Mostafa
  Rohaninejad, Nat McAleese, o3~contributors, Oleg M{\"u}rk, Rhythm Garg, Rui
  Shu, Szymon Sidor, Vineet Kosaraju, and Wenda Zhou.
\newblock Competitive programming with large reasoning models.
\newblock \emph{arXiv preprint arXiv:2502.06807}, 2025.

\bibitem[{Google DeepMind}(2025)]{gemini25}
{Google DeepMind}.
\newblock Gemini 2.5: Our newest gemini model with thinking.
\newblock
  \url{https://deepmind.google/blog/gemini-2-5-our-most-intelligent-ai-model/},
  2025.
\newblock Technical blog post.

\bibitem[Hubert et~al.(2025)Hubert, Mehta, Silver, et~al.]{alphaproof2025}
Thomas Hubert, Rishi Mehta, David Silver, et~al.
\newblock Olympiad-level formal mathematical reasoning with reinforcement
  learning.
\newblock \emph{Nature}, 2025.
\newblock \doi{10.1038/s41586-025-09833-y}.

\bibitem[{IOI}(2026)]{ioi}
{IOI}.
\newblock {International Olympiad in Informatics (IOI)}.
\newblock \url{https://www.ioinformatics.org/}, 2026.
\newblock Accessed 2026-03-25.

\bibitem[Jiang et~al.(2025)Jiang, Cai, Tian, Jia, Wang, and Wu]{dcp2025}
Chenyu Jiang, Zhenkun Cai, Ye~Tian, Zhen Jia, Yida Wang, and Chuan Wu.
\newblock {DCP}: Addressing input dynamism in long-context training via dynamic
  context parallelism.
\newblock In \emph{Proceedings of the ACM SIGOPS 31st Symposium on Operating
  Systems Principles}, pp.\  221--236, 2025.

\bibitem[Kuroki et~al.(2024)Kuroki, Nakamura, Akiba, and Tang]{cycleqd2025}
So~Kuroki, Taishi Nakamura, Takuya Akiba, and Yujin Tang.
\newblock Agent skill acquisition for large language models via {CycleQD}.
\newblock \emph{arXiv preprint arXiv:2410.14735}, 2024.
\newblock ICLR 2025.

\bibitem[{LeetCode}(2026)]{leetcode}
{LeetCode}.
\newblock {LeetCode}.
\newblock \url{https://leetcode.com/}, 2026.
\newblock Accessed 2026-03-25.

\bibitem[Li et~al.(2023)Li, Jin, Liu, Lyu, Sun, Huang, Zhang, Fu, and
  Li]{taco2023}
Ge~Li, Zhi Jin, Guang Liu, Chen Lyu, Zhihong Sun, Tao Huang, Bo-Wen Zhang, Jie
  Fu, and Rongao Li.
\newblock {TACO}: Topics in algorithmic code generation dataset.
\newblock \emph{arXiv preprint arXiv:2312.14852}, 2023.

\bibitem[Li et~al.(2025{\natexlab{a}})Li, Li, Dong, Zhang, Ruan, Dai, Liu, Xu,
  Wang, and Tang]{hlce2025}
Xiangyang Li, Xiaopeng Li, Kuicai Dong, Quanhu Zhang, Rongju Ruan, Xinyi Dai,
  Xiaoshuang Liu, Shengchun Xu, Yasheng Wang, and Ruiming Tang.
\newblock Humanity's last code exam: Can advanced {LLMs} conquer human's
  hardest code competition?
\newblock \emph{arXiv preprint arXiv:2506.12713}, 2025{\natexlab{a}}.

\bibitem[Li et~al.(2025{\natexlab{b}})Li, Sun, Wang, Li, and Shum]{li2025cuda}
Xiaoya Li, Xiaofei Sun, Albert Wang, Jiwei Li, and Chris Shum.
\newblock {Cuda-l1}: Improving {CUDA} optimization via contrastive
  reinforcement learning.
\newblock \emph{arXiv preprint arXiv:2507.14111}, 2025{\natexlab{b}}.

\bibitem[Li et~al.(2022)Li, Choi, Chung, Kushman, Schrittwieser, Leblond,
  Eccles, Keeling, Gimeno, Dal~Lago, et~al.]{li2022alphacode}
Yujia Li, David Choi, Junyoung Chung, Nate Kushman, Julian Schrittwieser,
  R{\'e}mi Leblond, Tom Eccles, James Keeling, Felix Gimeno, Agustin Dal~Lago,
  et~al.
\newblock Competition-level code generation with {AlphaCode}.
\newblock \emph{Science}, 378\penalty0 (6624):\penalty0 1092--1097, 2022.

\bibitem[Liu et~al.(2025)Liu, Zhang, Zhu, Dong, Zhou, Shang, Yang, and
  Yang]{rstarcoder2025}
Yifei Liu, Li~Lyna Zhang, Yi~Zhu, Bingcheng Dong, Xudong Zhou, Ning Shang, Fan
  Yang, and Mao Yang.
\newblock {rStar-Coder}: Scaling competitive code reasoning with a large-scale
  verified dataset.
\newblock \emph{arXiv preprint arXiv:2505.21297}, 2025.

\bibitem[{Llama Team, AI @ Meta}(2024)]{llama3}
{Llama Team, AI @ Meta}.
\newblock The {Llama} 3 herd of models.
\newblock \emph{arXiv preprint arXiv:2407.21783}, 2024.

\bibitem[Ma et~al.(2026)Ma, Zhang, Zhang, Yang, Zhang, Zhang, Jing, Zhang,
  Zheng, Zhao, Lin, and Hui]{agenticverifier2026}
Zeyao Ma, Jing Zhang, Xiaokang Zhang, Jiaxi Yang, Zongmeng Zhang, Jiajun Zhang,
  Yuheng Jing, Lei Zhang, Hao Zheng, Wenting Zhao, Junyang Lin, and Binyuan
  Hui.
\newblock Scaling agentic verifier for competitive coding.
\newblock \emph{arXiv preprint arXiv:2602.04254}, 2026.

\bibitem[{Moonshot AI}(2025)]{kimi25}
{Moonshot AI}.
\newblock Kimi k2.5 release.
\newblock \url{https://www.moonshot.ai/news/kimi-k2-5-release}, 2025.
\newblock Technical report.

\bibitem[OpenAI(2023)]{openai2023gpt4}
OpenAI.
\newblock {GPT-4} technical report.
\newblock \emph{arXiv preprint arXiv:2303.08774}, 2023.

\bibitem[OpenAI(2024{\natexlab{a}})]{openai2024gpt4o}
OpenAI.
\newblock {GPT-4o} system card.
\newblock \emph{arXiv preprint arXiv:2410.21276}, 2024{\natexlab{a}}.

\bibitem[OpenAI(2024{\natexlab{b}})]{openai2024o1}
OpenAI.
\newblock Openai o1 system card.
\newblock \emph{arXiv preprint arXiv:2412.16720}, 2024{\natexlab{b}}.

\bibitem[OpenAI(2025)]{openai2025o3}
OpenAI.
\newblock Openai o3 and o4-mini system card.
\newblock Technical report, 2025.
\newblock URL
  \url{https://cdn.openai.com/pdf/2221c875-02dc-4789-800b-e7758f3722c1/o3-and-o4-mini-system-card.pdf}.

\bibitem[Pich{\'e} et~al.(2025)Pich{\'e}, Kamalloo, Pardinas, Chen, and
  Bahdanau]{piche2025pipelinerl}
Alexandre Pich{\'e}, Ehsan Kamalloo, Rafael Pardinas, Xiaoyin Chen, and Dzmitry
  Bahdanau.
\newblock Pipelinerl: Faster on-policy reinforcement learning for long sequence
  generation.
\newblock \emph{arXiv preprint arXiv:2509.19128}, 2025.

\bibitem[Quan et~al.(2025)Quan, Yang, Yu, Zheng, Liu, Yang, Ren, Gao, Miao,
  Feng, Wang, Yang, Cui, Fan, Zhang, Hui, and Lin]{codeelo2025}
Shanghaoran Quan, Jiaxi Yang, Bowen Yu, Bo~Zheng, Dayiheng Liu, An~Yang,
  Xuancheng Ren, Bofei Gao, Yibo Miao, Yunlong Feng, Zekun Wang, Jian Yang,
  Zeyu Cui, Yang Fan, Yichang Zhang, Binyuan Hui, and Junyang Lin.
\newblock {CodeElo}: Benchmarking competition-level code generation of {LLMs}
  with human-comparable elo ratings.
\newblock \emph{arXiv preprint arXiv:2501.01257}, 2025.

\bibitem[Shao et~al.(2024)Shao, Wang, Zhu, Xu, Song, Bi, Zhang, Zhang, Li, Wu,
  and Guo]{deepseekmath2024}
Zhihong Shao, Peiyi Wang, Qihao Zhu, Runxin Xu, Junxiao Song, Xiao Bi, Haowei
  Zhang, Mingchuan Zhang, Y.K. Li, Y.~Wu, and Daya Guo.
\newblock {DeepSeekMath}: Pushing the limits of mathematical reasoning in open
  language models.
\newblock \emph{arXiv preprint arXiv:2402.03300}, 2024.

\bibitem[Surina et~al.(2025)Surina, Mansouri, Quaedvlieg, Seddas, Viazovska,
  Abbe, and Gulcehre]{algodiscovery2025}
Anja Surina, Amin Mansouri, Lars Quaedvlieg, Amal Seddas, Maryna Viazovska,
  Emmanuel Abbe, and Caglar Gulcehre.
\newblock Algorithm discovery with {LLM}s: Evolutionary search meets
  reinforcement learning.
\newblock \emph{arXiv preprint arXiv:2504.05108}, 2025.

\bibitem[{Thinking Machines Lab}(2025)]{tinker2025}
{Thinking Machines Lab}.
\newblock Tinker.
\newblock \url{https://github.com/thinking-machines-lab/tinker}, 2025.
\newblock Open-source training API.

\bibitem[{THUDM}(2026)]{slime2026}
{THUDM}.
\newblock slime.
\newblock \url{https://github.com/THUDM/slime}, 2026.
\newblock Open-source reinforcement learning post-training framework.

\bibitem[{USACO}(2026)]{usaco}
{USACO}.
\newblock {USA Computing Olympiad (USACO)}.
\newblock \url{http://www.usaco.org/}, 2026.
\newblock Accessed 2026-03-25.

\bibitem[Wang et~al.(2025{\natexlab{a}})Wang, Chen, Liu, Mak, Du, Moon, Xu,
  Tua, Peng, Lu, Xia, Zou, Ran, Tian, Zhu, Duan, Kang, Lin, Li, Luo, Long,
  Chen, Xiao, Wu, Zan, Fu, Wang, and Ding]{aethercode2025}
Zihan Wang, Jiaze Chen, Zhicheng Liu, Markus Mak, Yidi Du, Geonsik Moon, Luoqi
  Xu, Aaron Tua, Kunshuo Peng, Jiayi Lu, Mingfei Xia, Boqian Zou, Chenyang Ran,
  Guang Tian, Shoutai Zhu, Yeheng Duan, Zhenghui Kang, Zhenxing Lin, Shangshu
  Li, Qiang Luo, Qingshen Long, Zhiyong Chen, Yihan Xiao, Yurong Wu, Daoguang
  Zan, Yuyi Fu, Mingxuan Wang, and Ming Ding.
\newblock {AetherCode}: Evaluating {LLMs}' ability to win in premier
  programming competitions.
\newblock \emph{arXiv preprint arXiv:2508.16402}, 2025{\natexlab{a}}.

\bibitem[Wang et~al.(2025{\natexlab{b}})Wang, Liu, Sun, Li, and
  Shen]{codecontestsplus2025}
Zihan Wang, Siyao Liu, Yang Sun, Hongyan Li, and Kai Shen.
\newblock {CodeContests+}: High-quality test case generation for competitive
  programming.
\newblock \emph{arXiv preprint arXiv:2506.05817}, 2025{\natexlab{b}}.

\bibitem[Xia et~al.(2026)Xia, Chen, Wang, Liu, Zeng, Wang, Han, Zhou, Zhao,
  Chen, Zheng, Xie, and Yao]{skillrl2026}
Peng Xia, Jianwen Chen, Hanyang Wang, Jiaqi Liu, Kaide Zeng, Yu~Wang, Siwei
  Han, Yiyang Zhou, Xujiang Zhao, Haifeng Chen, Zeyu Zheng, Cihang Xie, and
  Huaxiu Yao.
\newblock {SkillRL}: Evolving agents via recursive skill-augmented
  reinforcement learning.
\newblock \emph{arXiv preprint arXiv:2602.08234}, 2026.

\bibitem[Yang et~al.(2024)Yang, Yang, Zhang, Hui, Zheng, Yu, Li, Liu, Huang,
  et~al.]{qwen25}
An~Yang, Baosong Yang, Beichen Zhang, Binyuan Hui, Bo~Zheng, Bowen Yu,
  Chengyuan Li, Dayiheng Liu, Fei Huang, et~al.
\newblock Qwen2.5 technical report.
\newblock \emph{arXiv preprint arXiv:2412.15115}, 2024.

\bibitem[Yang et~al.(2025)Yang, Jin, Shi, Peng, Chen, and Xiong]{probench2025}
Lei Yang, Renren Jin, Ling Shi, Jianxiang Peng, Yue Chen, and Deyi Xiong.
\newblock {ProBench}: Benchmarking large language models in competitive
  programming.
\newblock \emph{arXiv preprint arXiv:2502.20868}, 2025.

\bibitem[Yu et~al.(2025)Yu, Chen, Feng, Chen, Dai, Yu, Zhang, Ma, Liu, Wang,
  and Zhou]{memagent2025}
Hongli Yu, Tinghong Chen, Jiangtao Feng, Jiangjie Chen, Weinan Dai, Qiying Yu,
  Ya-Qin Zhang, Wei-Ying Ma, Jingjing Liu, Mingxuan Wang, and Hao Zhou.
\newblock {MemAgent}: Reshaping long-context {LLM} with multi-conv {RL}-based
  memory agent.
\newblock \emph{arXiv preprint arXiv:2507.02259}, 2025.

\bibitem[Yu et~al.(2026)Yu, Yao, Xie, Tan, Feng, Li, and Wu]{agenticmemory2026}
Yi~Yu, Liuyi Yao, Yuexiang Xie, Qingquan Tan, Jiaqi Feng, Yaliang Li, and
  Libing Wu.
\newblock Agentic memory: Learning unified long-term and short-term memory
  management for large language model agents.
\newblock \emph{arXiv preprint arXiv:2601.01885}, 2026.

\bibitem[Yuksekgonul et~al.(2026)Yuksekgonul, Koceja, Li, Bianchi, McCaleb,
  Wang, Kautz, Choi, Zou, Guestrin, and Sun]{tttdiscover2026}
Mert Yuksekgonul, Daniel Koceja, Xinhao Li, Federico Bianchi, Jed McCaleb,
  Xiaolong Wang, Jan Kautz, Yejin Choi, James Zou, Carlos Guestrin, and Yu~Sun.
\newblock Learning to discover at test time.
\newblock \emph{arXiv preprint arXiv:2601.16175}, 2026.

\bibitem[Zheng et~al.(2026)Zheng, Dong, Liu, Oliva, Yong, Lin, Chen, Wang, and
  Hassan]{whenelolies2026}
Shenyu Zheng, Ximing Dong, Xiaoshuang Liu, Gustavo Oliva, Chong~Chun Yong, Dayi
  Lin, Boyuan Chen, Shaowei Wang, and Ahmed~E. Hassan.
\newblock When elo lies: Hidden biases in codeforces-based evaluation of large
  language models.
\newblock \emph{arXiv preprint arXiv:2602.05891}, 2026.

\bibitem[{Zhipu AI and Tsinghua University KEG}(2024)]{glm4}
{Zhipu AI and Tsinghua University KEG}.
\newblock {GLM}-4: Open multilingual multimodal chat lms.
\newblock \url{https://github.com/zai-org/GLM-4}, 2024.
\newblock Project page.

\end{thebibliography}
\bibliographystyle{colm}
\clearpage
\appendix
\section{Contributors}
Xiaoya Li, Guoyin Wang, Songqiao Su, Jiwei Li, Chris Shum
 
{\{xiaoya\_li, songqiao\_su, jiwei\_li, chris\_shum\}@ornith.ai}
\section{Submission Details}

\begin{figure}
\centering
\begin{subfigure}[t]{\linewidth}
\centering
\begin{minipage}{0.49\linewidth}
\centering
\includegraphics[width=0.9215\linewidth]{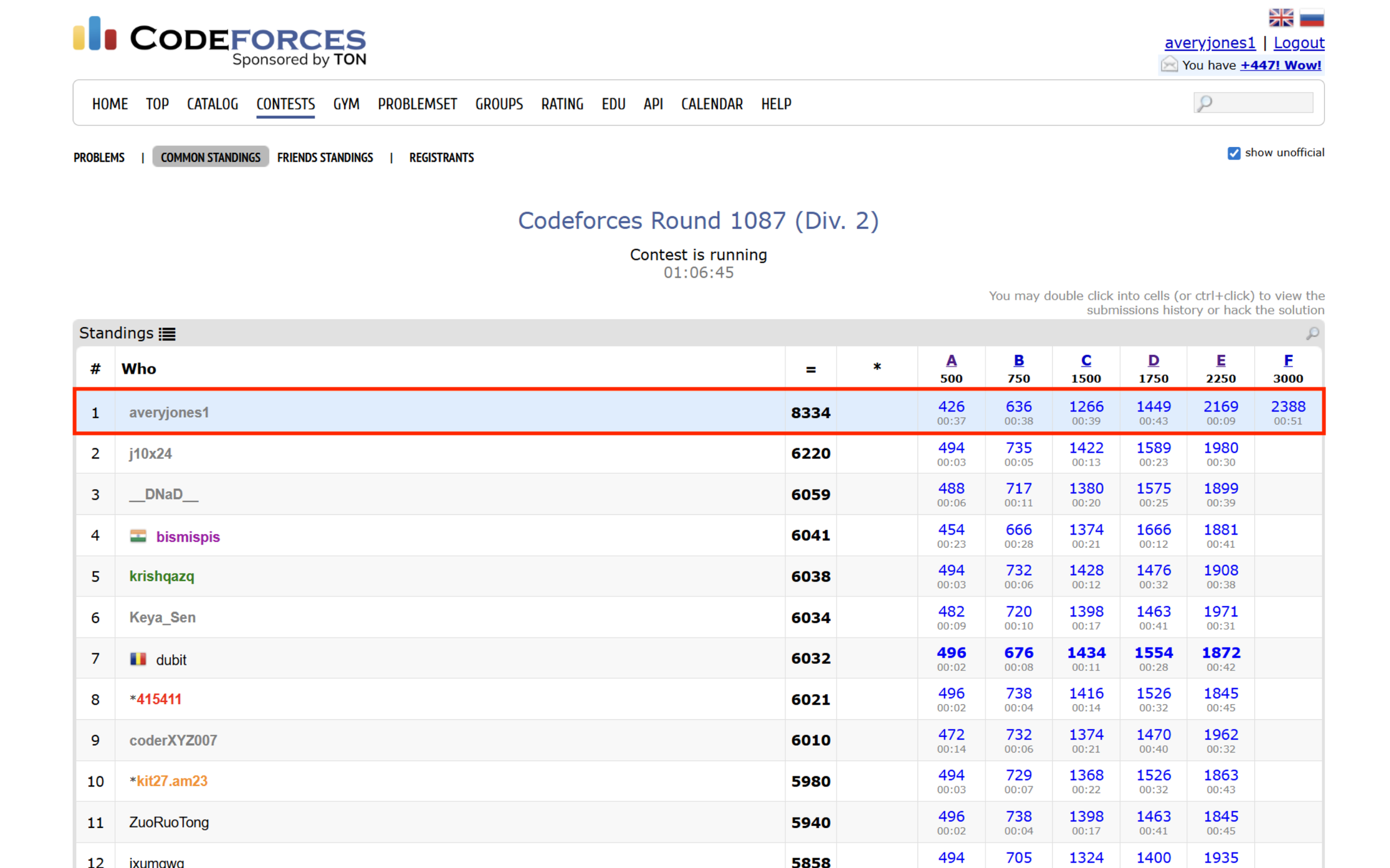}
\end{minipage}
\hfill
\begin{minipage}{0.49\linewidth}
\centering
\includegraphics[width=1.0815\linewidth]{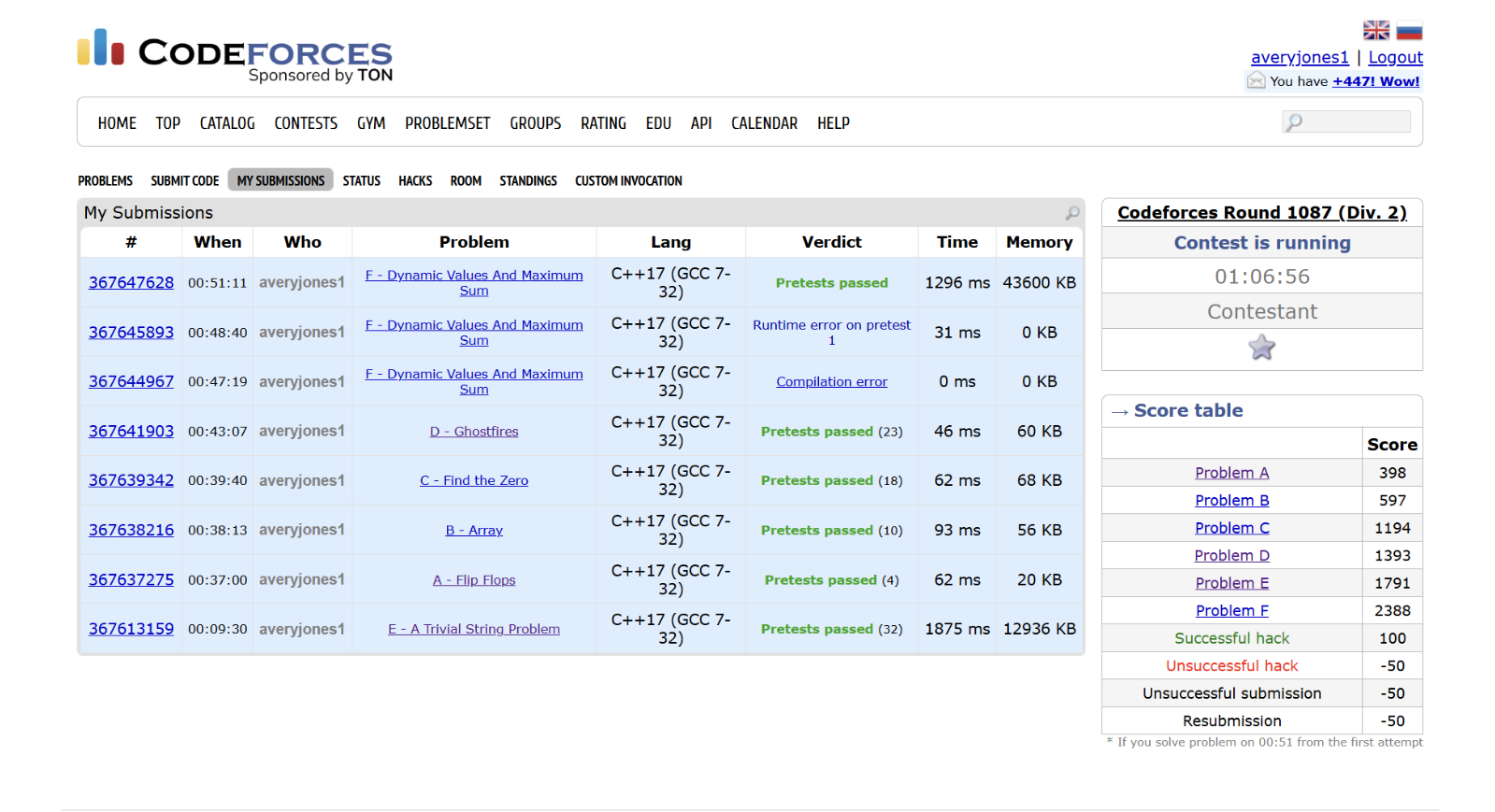}
\end{minipage}
\caption{Round~1087, ID \texttt{averyjones1}.}
\label{fig:submission-1087}
\end{subfigure}

\vspace{0.8em}
\begin{subfigure}[t]{\linewidth}
\centering
\begin{minipage}{0.49\linewidth}
\centering
\includegraphics[width=0.9215\linewidth]{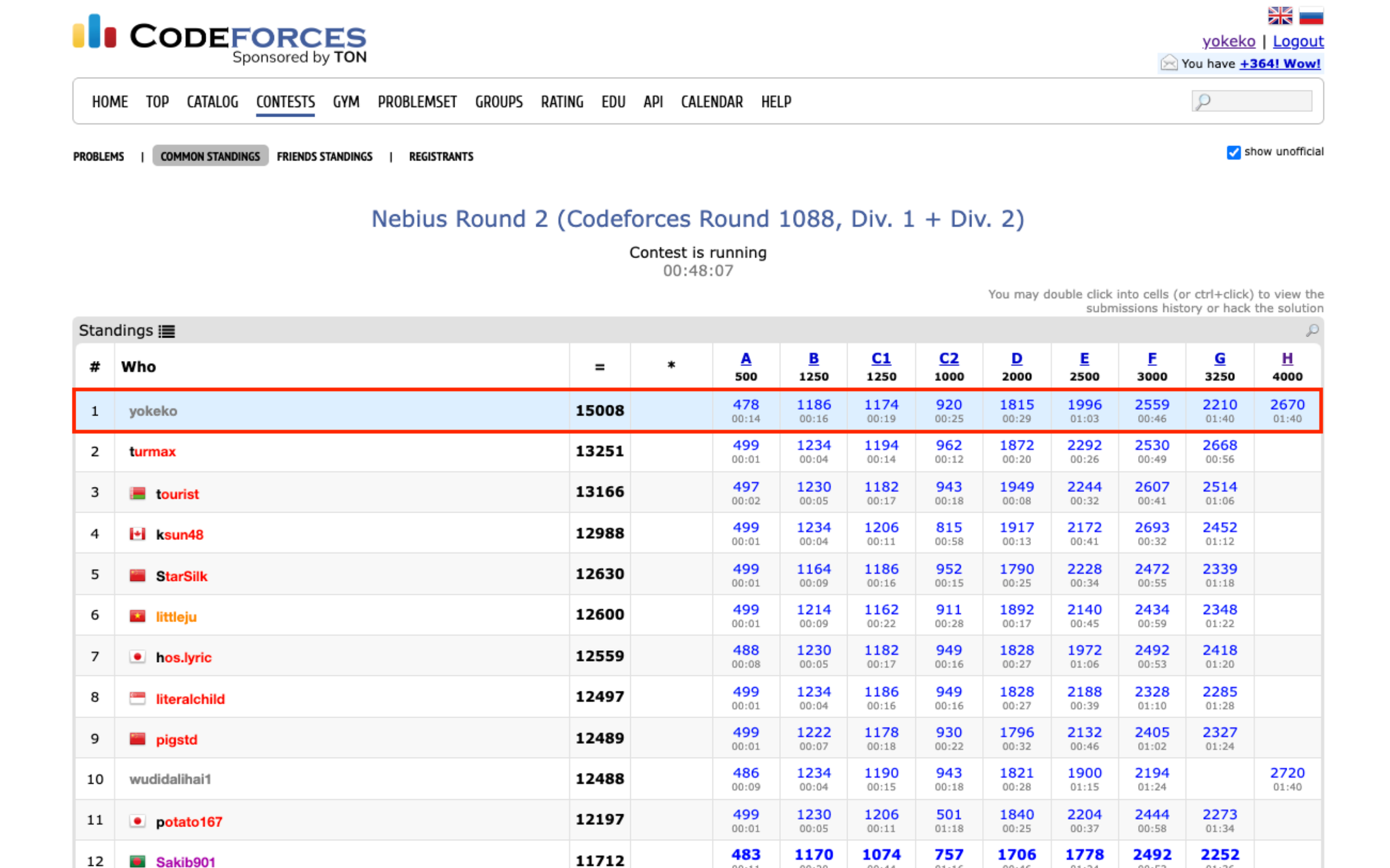}
\end{minipage}
\hfill
\begin{minipage}{0.49\linewidth}
\centering
\includegraphics[width=1.0815\linewidth]{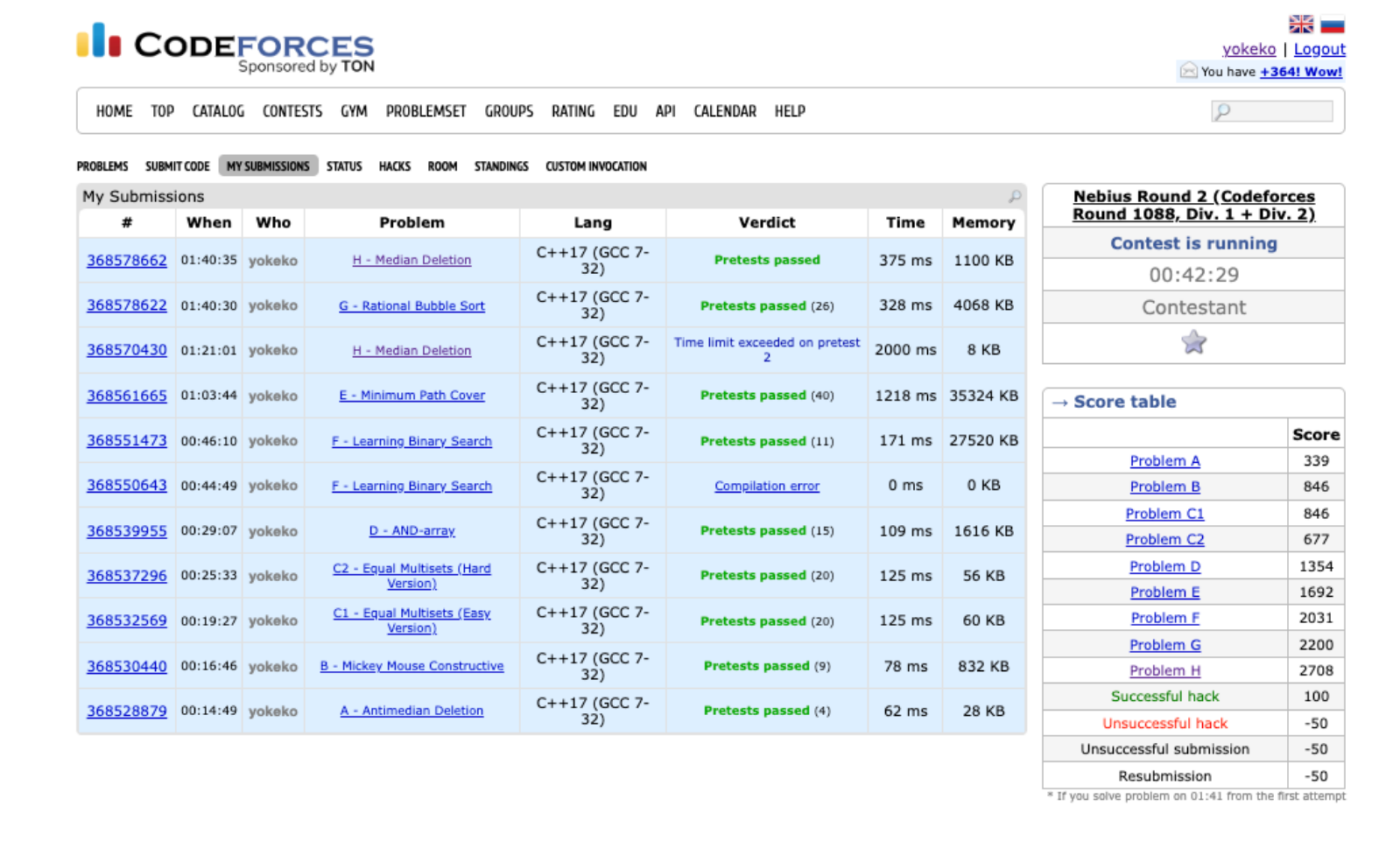}
\end{minipage}
\caption{Round~1088, ID \texttt{yokeko}.}
\label{fig:submission-1088}
\end{subfigure}

\vspace{0.8em}
\begin{subfigure}[t]{\linewidth}
\centering
\begin{minipage}{0.49\linewidth}
\centering
\includegraphics[width=0.9215\linewidth]{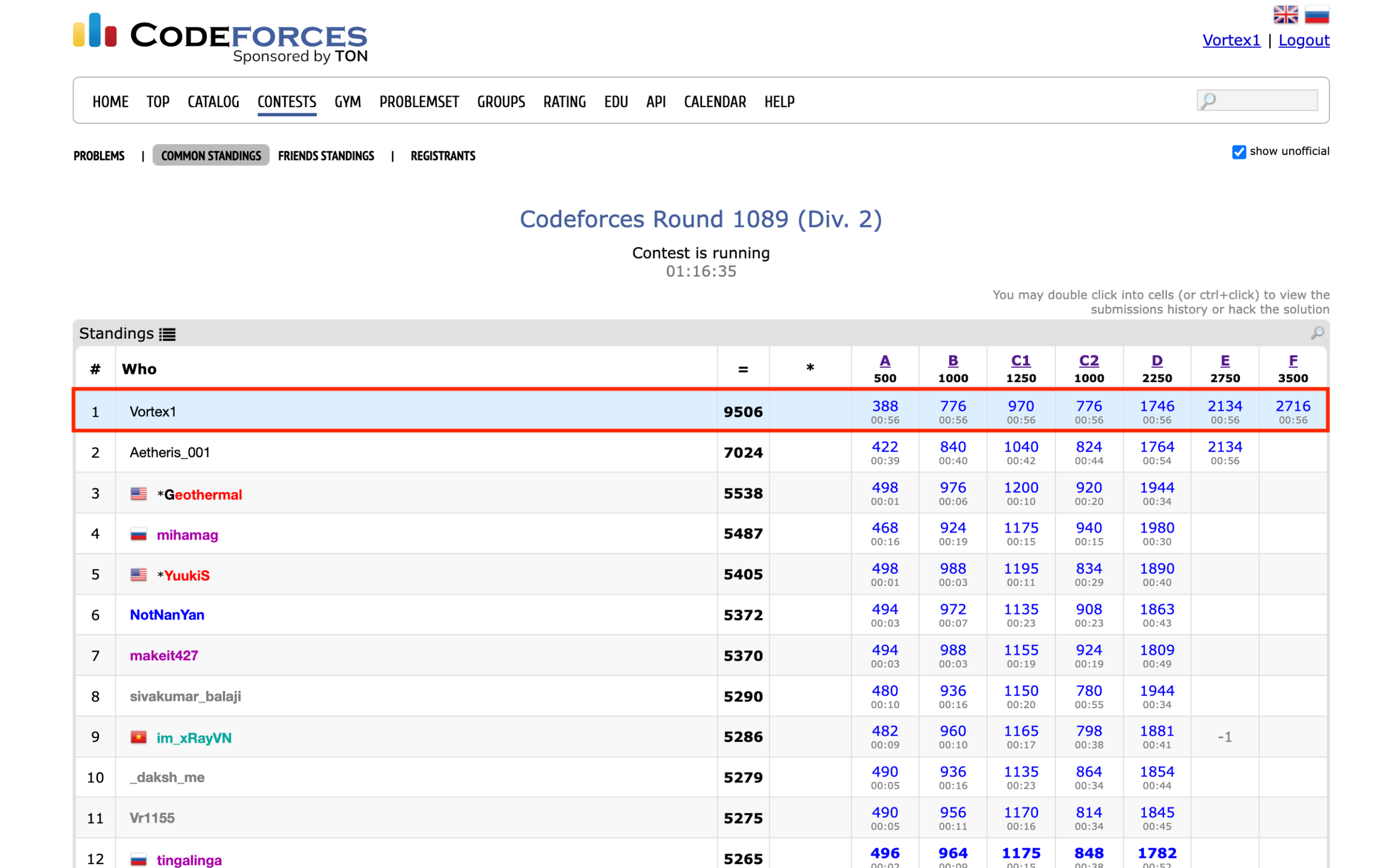}
\end{minipage}
\hfill
\begin{minipage}{0.49\linewidth}
\centering
\includegraphics[width=1.0815\linewidth]{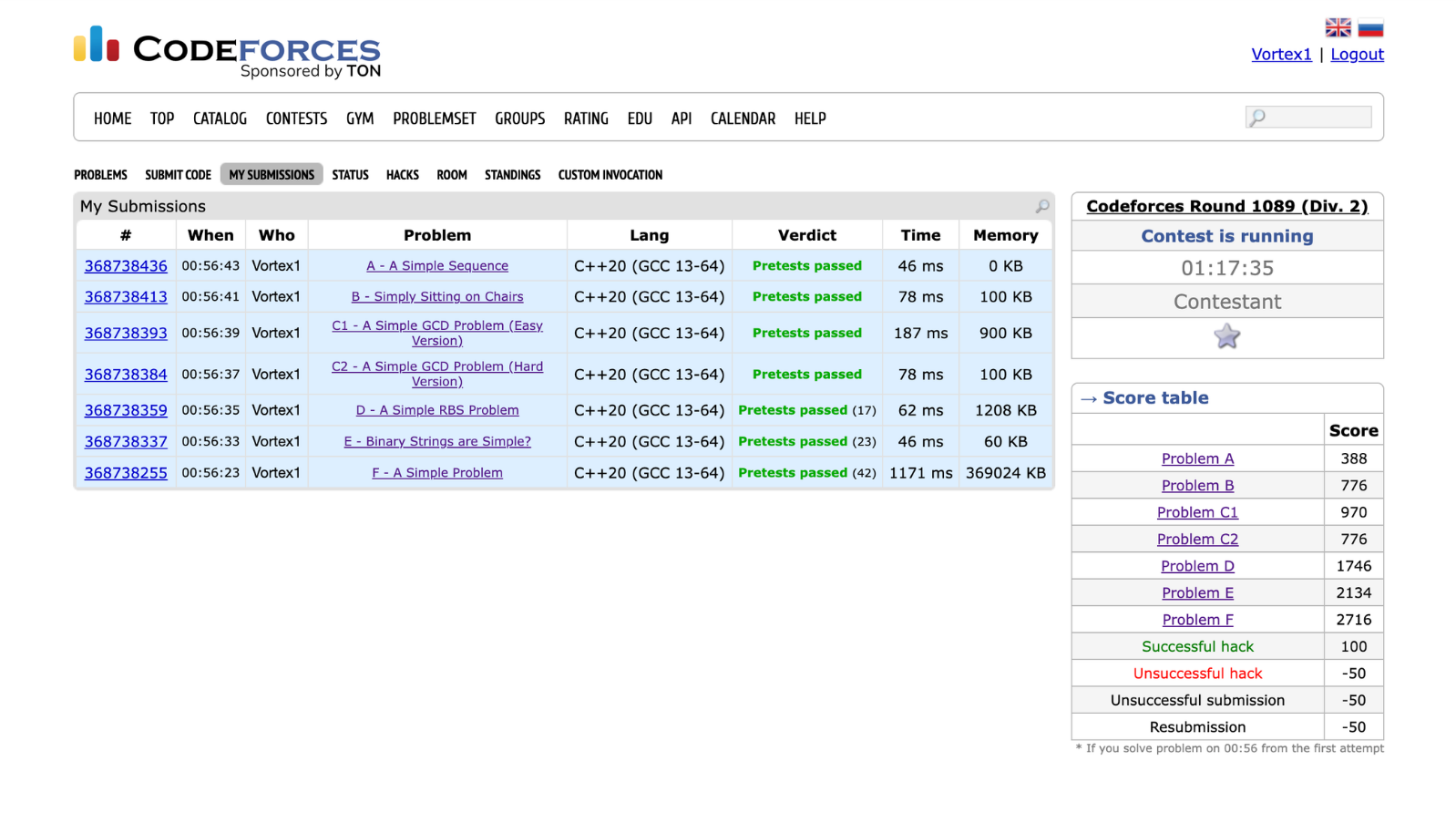}
\end{minipage}
\caption{Round~1089, ID \texttt{Vortex1}.}
\label{fig:submission-1089}
\end{subfigure}
\caption{Standings and submission pages for GrandCode in the three live
Codeforces contests. The score corresponds to 
 $S(\mathrm{joint})$, which is
based on the full set of submissions in a single account.}
\label{fig:submission-details}
\end{figure}

Figure~\ref{fig:submission-details} shows the standings pages together with the corresponding
submission pages 
for the {\it joint} setup
 in the three live Codeforces contests, where all codes need to be submitted in a single account. 
These screenshots provide direct evidence of the contest identities,
submission accounts, scores, accepted solutions, and the times at which
the full problem sets were completed. The three paired screenshots
correspond to Round~1087 under the ID \texttt{averyjones1},
Round~1088 under the ID \texttt{yokeko}, and Round~1089 under the ID
\texttt{Vortex1}.

\section{Code Reward}
\label{CodeReward}
Reward for a generated code snippet can summarized formally as follows. Let
$P$ denote a generated code snippet, and let
$T = \{t_1, \ldots, t_m\}$ denote the generated test set. For each test
case $t_i$, define the reference output
\begin{equation}
r_i =
\begin{cases}
y^\star(t_i), & \text{if a gold solution } y^\star \text{ is available}, \\
b(t_i), & \text{otherwise, where } b \text{ is a brute-force solver.}
\end{cases}
\end{equation}
Define the executability and correctness indicators by
\begin{equation}
E(P) =
\begin{cases}
1, & \text{if } P \text{ compiles and executes}, \\
0, & \text{otherwise,}
\end{cases}
\qquad
C(P) =
\begin{cases}
1, & \text{if } P(t_i) = r_i \text{ for all correctness-check tests}, \\
0, & \text{otherwise.}
\end{cases}
\end{equation}
For each efficiency test case $t_i$, define
\begin{equation}
s_i(P) =
\begin{cases}
\dfrac{\tau_b(t_i)}{\tau_P(t_i)}, & \text{if } P \text{ finishes within the time limit}, \\
0.1, & \text{if } P \text{ times out,}
\end{cases}
\end{equation}
where $\tau_b(t_i)$ and $\tau_P(t_i)$ denote the runtimes of the
brute-force solver and the generated code snippet, respectively. The final
score is then
\begin{equation}
R(P) =
\begin{cases}
0, & \text{if } E(P) = 0, \\
0, & \text{if } C(P) = 0, \\
\dfrac{1}{m} \sum_{i=1}^{m} s_i(P), & \text{otherwise.}
\end{cases}
\end{equation}

\begin{figure}[p]
\centering
\begin{minipage}{0.45\linewidth}
\centering
\includegraphics[width=\linewidth,height=\textheight,keepaspectratio]{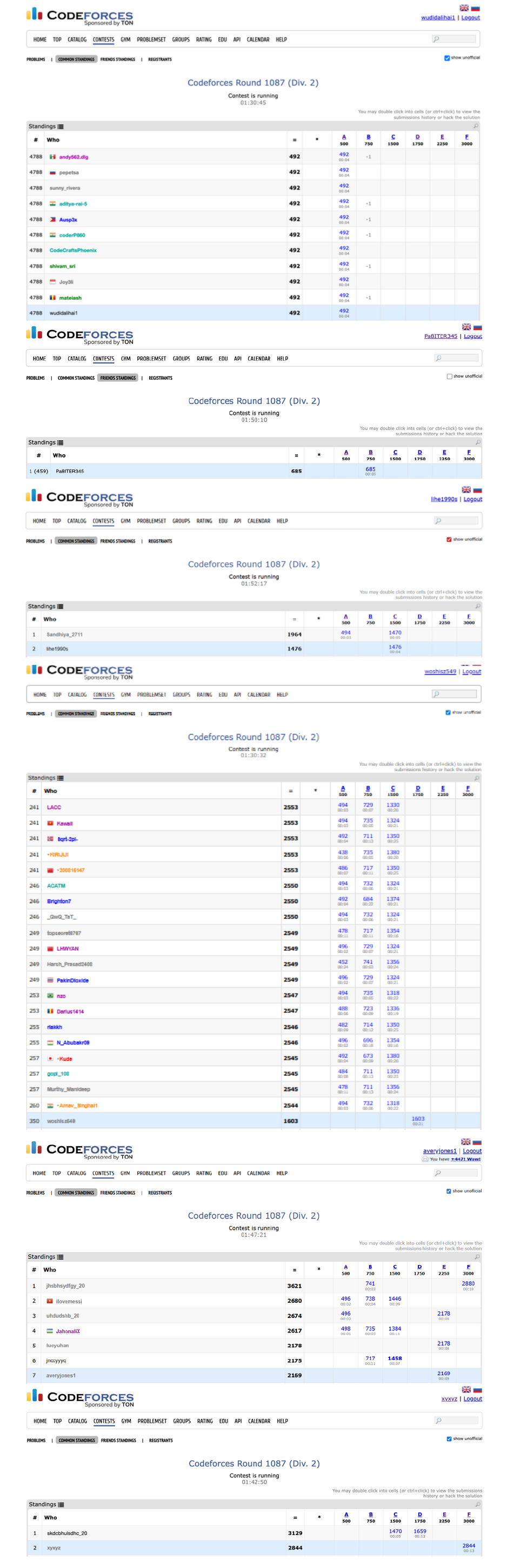}
\end{minipage}
\hfill
\begin{minipage}{0.52\linewidth}
\centering
\includegraphics[width=\linewidth,height=\textheight,keepaspectratio]{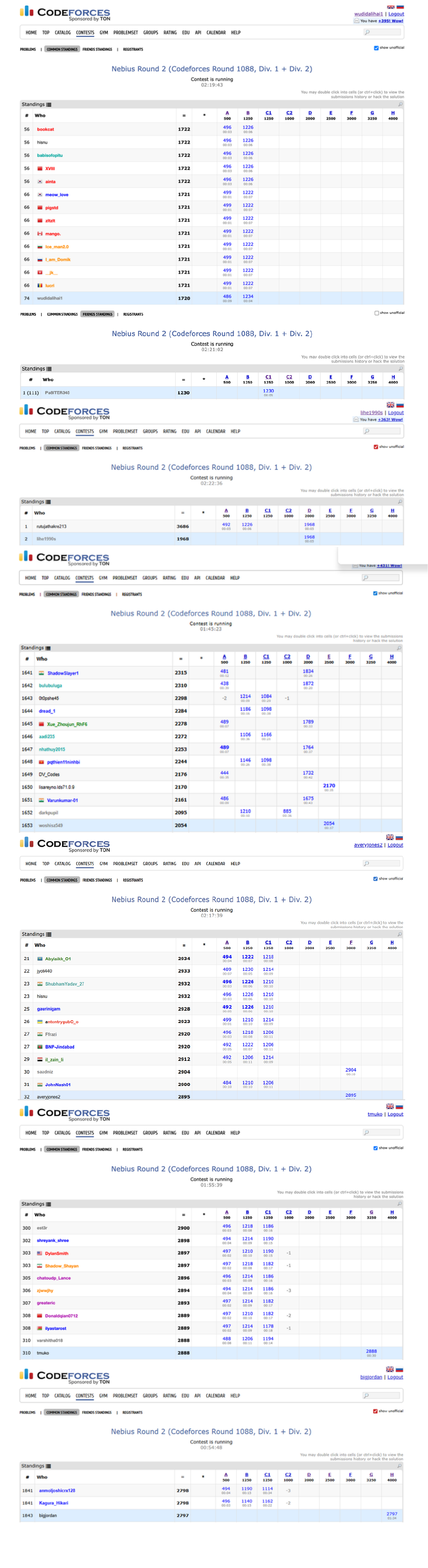}
\end{minipage}
\caption{Standings and submission pages for GrandCode. The score corresponds to $S(\mathrm{seperate})$, the score achieved by each solution as soon as it is ready.
(a)~Round~1087. A--492, B--685, C--1476, D--1603, E--2169, F--2844; $S(\mathrm{seperate})$: 9269.
(b)~Round~1088. A--486, B--1234, C1--1230, C2--959, D--1968, E--2054, F--2895, G--2888, H--2797; $S(\mathrm{seperate})$: 16511.}
\label{fig:ind-scores-1087-1088}
\end{figure}

\begin{figure}[p]
\floatpagestyle{empty}
\centering
\includegraphics[width=\linewidth,height=\textheight,keepaspectratio]{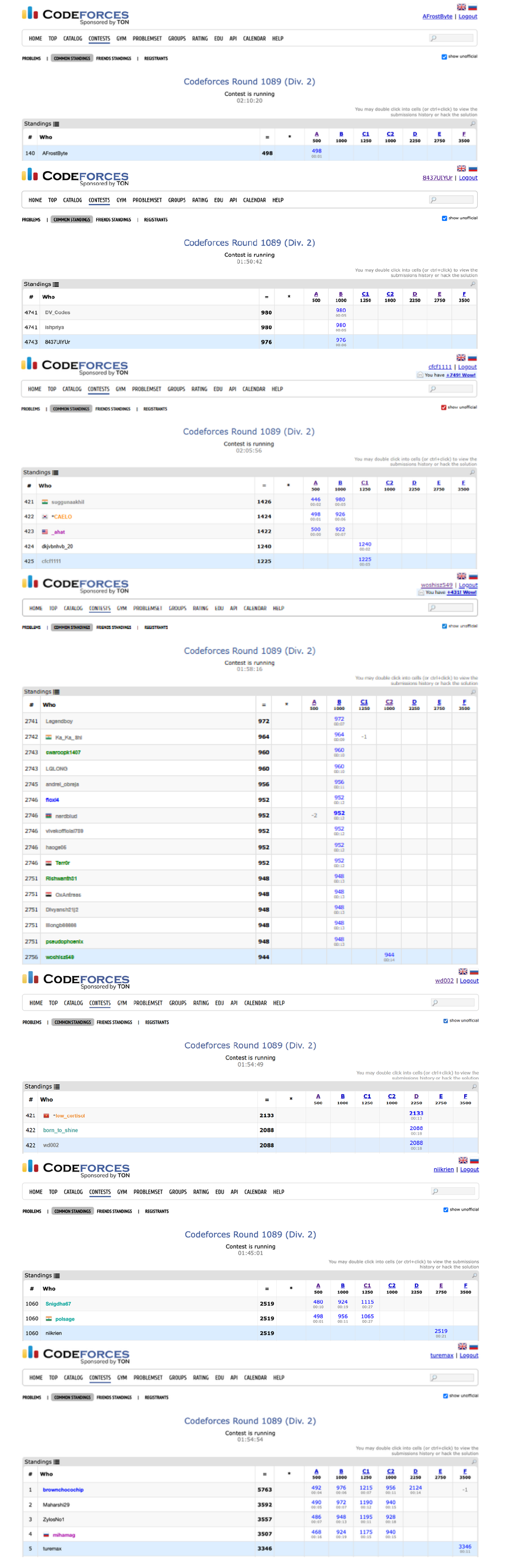}
\caption{Standings and submission pages for GrandCode in Round~1089.
The score corresponds to $S(\mathrm{seperate})$, the score achieved by
each solution as soon as it is ready.
A\,--\,498,
B\,--\,976,
C1\,--\,1225,
C2\,--\,944,
D\,--\,2088,
E\,--\,2519,
F\,--\,3346;
$S(\mathrm{seperate})$: 11596.}
\label{fig:ind-scores-1089}
\end{figure}

\begin{figure}[p]
\centering
\begin{tcolorbox}[
  colback=white, colframe=black!70,
  boxsep=1pt, left=4pt, right=4pt, top=3pt, bottom=3pt,
  arc=1pt, boxrule=0.5pt,
  width=\linewidth,
]
{\scriptsize
\textbf{G.\ Toothless}\hfill
Time limit: 2\,s\enspace$\mid$\enspace Memory: 256\,MB\enspace$\mid$\enspace
I/O: standard input / standard output\par
\smallskip
Let $n, m, a, b$ be positive integers with $a \le b$.
Toothless is drawing in an $n \times m$ grid of sand that is initially all white.
In one move, he may do the following:
\begin{itemize}[leftmargin=1.5em, nosep]
\item Select any currently white cell and color it black if it has at most one
  black cell among its edge-adjacent neighbors.
\end{itemize}
Because he is particular about his art, Toothless thinks some cells in the grid
are \emph{needy}, and these cells must end up black.
Of the cells that aren't needy, he also thinks some of them are \emph{special}.
A cell has value $b$ if it is special, and $a$ otherwise.
\textbf{No special cell borders a needy cell by an edge.}

Let $P$ be the sum of the values of all the cells in the grid that
\textbf{aren't} needy.
Because Toothless is very particular about his art, he wants to make some number
of moves such that:
\begin{itemize}[leftmargin=1.5em, nosep]
\item Each of the needy cells becomes black after all moves are done,
\item The total value of cells that are colored black (\textbf{including}
  needy cells) is at least $\frac{2}{3} \cdot P$.
\end{itemize}
Compute any sequence of moves that Toothless could make.
Tests are generated in a such way, that it is guaranteed that all of the
needy cells in the input can be shaded black after some number of moves.
Furthermore, it can be shown that for the given constraints, a satisfying
sequence of moves always exists.\par
\smallskip
\textbf{Input.}\enspace
Each test contains multiple test cases. The first line contains the number of
test cases $t$ ($1 \le t \le 10^4$). The description of the test cases follows.

The first line of each test case contains four integers $n$, $m$, $a$, and $b$
($1 \le n, m \le 2 \cdot 10^3$,
$1 \le n \cdot m \le 2 \cdot 10^3$,
$1 \le a \le b \le 5 \cdot 10^5$)
--- the dimensions of the grid and the values of non-special and special cells,
respectively.
The $i$-th of the next $n$ lines each contain a string $s_i$ --- a string of
exactly $m$ characters depicting the $i$-th row of cells.
\begin{itemize}[leftmargin=1.5em, nosep]
\item The $j$-th character is `\texttt{\#}' if the cell at $(i,j)$ is needy.
\item The $j$-th character is `\texttt{x}' if the cell at $(i,j)$ is special.
\item Otherwise, the $j$-th character is `\texttt{.}'.
\end{itemize}
It is guaranteed that no needy cell is adjacent to a special cell.
Tests are generated in a such way, that it is guaranteed that all of the needy
cells in the input can be shaded black after some number of moves.
It is guaranteed that the sum of $(nm)^2$ over all test cases does not exceed
$4 \cdot 10^6$.\par
\smallskip
\textbf{Output.}\enspace
For each test case, print the following.
In the first line, print the number of moves $\mathit{op}$
($0 \le \mathit{op} \le n \cdot m$) you want to make.
For each of the $\mathit{op}$ moves, print a line containing two integers
$i$, $j$ ($1 \le i \le n$, $1 \le j \le m$), indicating that you wish to
color the cell at $(i, j)$ black.
}%
\tcblower
\begin{tcolorbox}[
  colback=gray!5, colframe=black!40,
  boxsep=1pt, left=2pt, right=2pt, top=1pt, bottom=1pt,
  arc=0.5pt, boxrule=0.4pt,
  sidebyside, sidebyside align=top,
  lefthand width=0.35\linewidth,
  lower separated=false,
]
{\scriptsize\textbf{Input}}
\begin{lstlisting}[basicstyle=\ttfamily\fontsize{6}{7.2}\selectfont,
  numbers=none, frame=none, xleftmargin=0pt, aboveskip=1pt, belowskip=0pt]
3
3 3 1 5
#..
...
..x
2 3 1 2
...
xxx
3 5 8 9
x.x.x
.x.x.
x.#.x
\end{lstlisting}
\tcblower
{\scriptsize\textbf{Output}}
\begin{lstlisting}[basicstyle=\ttfamily\fontsize{6}{7.2}\selectfont,
  numbers=none, frame=none, xleftmargin=0pt, aboveskip=1pt, belowskip=0pt]
6
1 1
3 1
3 2
3 3
2 3
1 3
3
2 1
2 2
2 3
10
1 1
1 2
1 3
2 2
2 5
1 5
3 5
2 4
3 3
3 1
\end{lstlisting}
\end{tcolorbox}

\smallskip
{\scriptsize\textbf{Note.}\enspace
In the first test case, there is a needy cell in the top left corner and a
special cell in the bottom right corner.
Of the cells that aren't needy, there are 7 non-special cells worth $1$ each
and 1 special cell worth $5$, so $P = 12$.
So, we need to shade cells with a total value of at least
$\tfrac{2}{3} \cdot 12 = 8$;
the given sequence of moves achieves a value of $10$.
\enspace
In the second test case, there are no needy cells, but there are three special
cells in the bottom row of cells.
There are 3 non-special cells worth $1$ each and 3 special cells worth $2$, so
$P = 9$.
So, we need to shade cells with a total value of at least
$\tfrac{2}{3} \cdot 9 = 6$;
the given sequence of moves achieves a value of $6$.
\enspace
In the third test case, there is one needy cell at $(3,3)$, and $7$ special
cells.
Of the cells that aren't needy, there are 7 non-special cells worth $8$ each
and 7 special cells worth $9$, so $P = 7 \cdot 8 + 7 \cdot 9 = 119$.
So, we need to shade cells with a total value of at least
$\tfrac{2}{3} \cdot 119 = 79\tfrac{1}{3}$;
the given sequence of moves achieves a value of $87$.}

\medskip
\begin{minipage}[t]{0.30\linewidth}\centering
  \includegraphics[width=\linewidth]{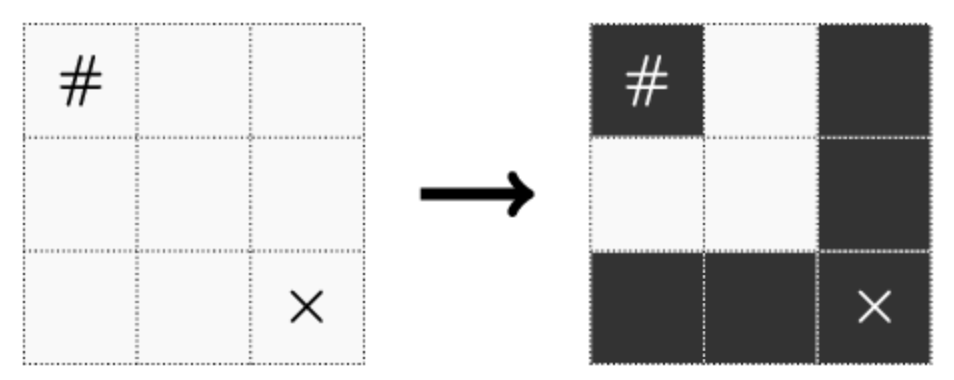}\\[-2pt]
  {\scriptsize (a) Test 1}
\end{minipage}\hfill
\begin{minipage}[t]{0.28\linewidth}\centering
  \includegraphics[width=\linewidth]{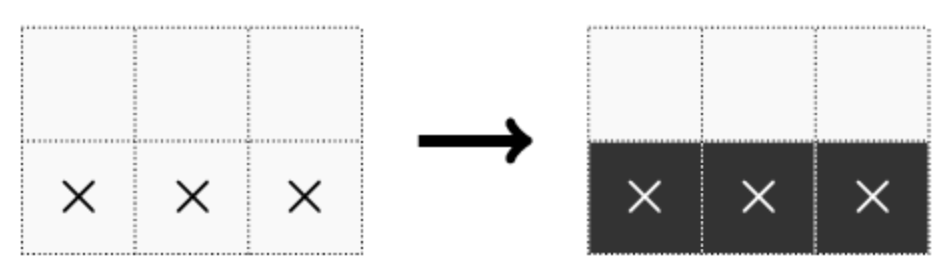}\\[-2pt]
  {\scriptsize (b) Test 2}
\end{minipage}\hfill
\begin{minipage}[t]{0.38\linewidth}\centering
  \includegraphics[width=\linewidth]{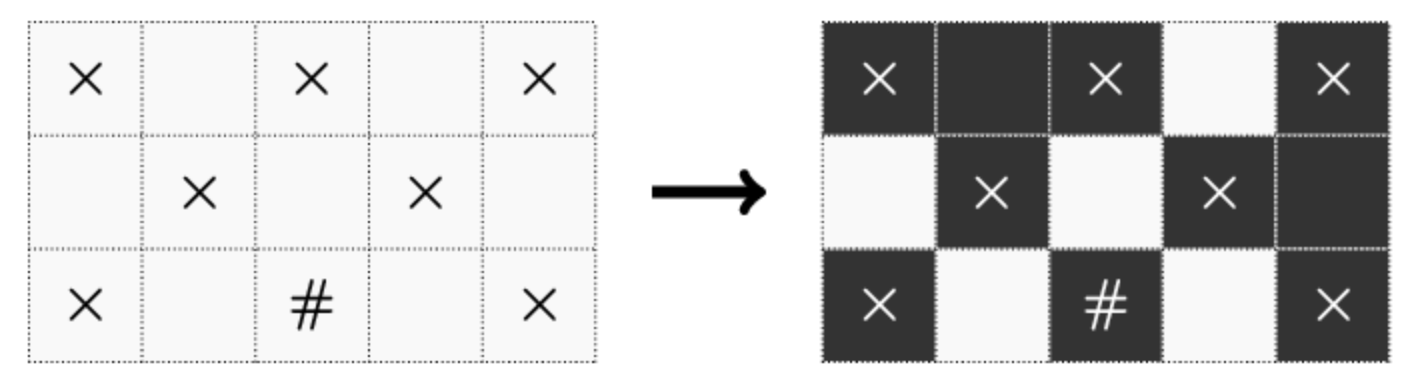}\\[-2pt]
  {\scriptsize (c) Test 3}
\end{minipage}

\end{tcolorbox}
\caption{Problem G.\ Toothless (Round~1088, Div.~1+2): complete
  statement, sample I/O, and example illustrations.
  Each panel shows the grid before (left) and after (right) moves;
  dark cells are colored black.}
\label{fig:cf-example}
\label{tab:cf-full-problem}
\end{figure}

\newpage
\section{Submitted Code}

\subsection*{Round~1089-A}
\begin{lstlisting}[style=cppcontest]
#include <iostream>
#include <vector>
#include <algorithm>
using namespace std;

int main() {
    ios::sync_with_stdio(false);
    cin.tie(nullptr);

    int a;
    cin >> a;
    while (a--) {
        int b;
        cin >> b;
        vector<int> c(b + 1), d(b + 1);
        for (int i = 1; i <= b; i++) {
            cin >> c[i];
            d[c[i]] = i;
        }

        int e = 0;
        int f = 0;
        int g = 1;
        while (g <= b) {
            if (d[g] < g) e++;
            int h = g - e;
            if (h > f) f = h;
            g++;
        }

        cout << f << '\n';
    }

    return 0;
}
\end{lstlisting}

\subsection*{Round~1088-E}
\begin{lstlisting}[style=cppcontest]
#include <bits/stdc++.h>
using namespace std;
typedef unsigned long long u64;

#if defined(_WIN32)||defined(_WIN64)
#define getchar_unlocked getchar
#endif

struct R {
    template<class T>
    bool read(T& x) {
        int c = getchar_unlocked();
        if (c == EOF) return false;
        while (c <= ' ') {
            c = getchar_unlocked();
            if (c == EOF) return false;
        }
        T v = 0;
        while (c > ' ') {
            v = v * 10 + (T)(c - '0');
            c = getchar_unlocked();
        }
        x = v;
        return true;
    }
};

struct S {
    static const int INF = 1e9;

    struct Fr {
        int a;
        u64 b;
        int c;
        int d;
        bool e;
    };

    int n;
    vector<u64> a;
    vector<vector<int>> b;
    vector<int> c;
    vector<int> d;
    vector<vector<pair<u64,int>>> e;
    vector<Fr> f;

    S(int n) : n(n), a(n+1,0), b(n+1), c(n+1,0), d(n+1,0), e(n+1) {}

    int fc(int u, u64 g) {
        auto& v = e[u];
        for (int i = 0; i < (int)v.size(); i++)
            if (v[i].first == g) return i;
        return -1;
    }

    int qr(int s, u64 g0) {
        if (g0 == 1) return INF;
        int p = fc(s, g0);
        if (p != -1) return e[s][p].second;

        f.clear();
        f.push_back({s, g0, 0, 0, false});

        while (!f.empty()) {
            Fr& cur = f.back();
            int cc = fc(cur.a, cur.b);
            if (cc != -1) { f.pop_back(); continue; }

            if (!cur.e) {
                if (gcd(a[cur.a], cur.b) == 1ULL) {
                    e[cur.a].push_back({cur.b, INF});
                    f.pop_back();
                    continue;
                }
                cur.e = true;
                cur.c = 0;
                cur.d = 0;
            }

            bool pushed = false;
            auto& ch = b[cur.a];
            while (cur.c < (int)ch.size()) {
                int v = ch[cur.c];
                u64 ng = gcd(cur.b, a[v]);
                if (ng == 1ULL) { cur.c++; continue; }
                int ci = fc(v, ng);
                if (ci == -1) {
                    f.push_back({v, ng, 0, 0, false});
                    pushed = true;
                    break;
                }
                cur.d = min(cur.d, e[v][ci].second - c[v] - 1);
                cur.c++;
            }
            if (pushed) continue;

            e[cur.a].push_back({cur.b, d[cur.a] + cur.d});
            f.pop_back();
        }

        p = fc(s, g0);
        return e[s][p].second;
    }
};

int main() {
    R in;
    int t;
    in.read(t);

    while (t--) {
        int n;
        in.read(n);

        S sv(n);
        vector<int> ind(n+1, 0);
        long long ec = 0;
        bool bad = false;

        for (int u = n; u >= 1; u--) {
            int k;
            in.read(sv.a[u]);
            in.read(k);
            ec += k;

            sv.b[u].reserve(k);
            int bs = 1;

            for (int i = 0; i < k; i++) {
                int v;
                in.read(v);
                sv.b[u].push_back(v);
                if (v <= u || v > n) bad = true;
                else {
                    ind[v]++;
                    if (ind[v] > 1) bad = true;
                }
                bs += sv.c[v];
            }

            sv.d[u] = bs;
            sv.e[u].reserve(4);
            sv.c[u] = sv.qr(u, sv.a[u]);

            printf("%d\n", sv.c[u]);
            fflush(stdout);
        }

        if (ec != n-1) bad = true;
        if (ind[1] != 0) bad = true;
        for (int i = 2; i <= n; i++) {
            if (ind[i] != 1) { bad = true; break; }
        }

        if (bad) {
            printf("-1\n");
            fflush(stdout);
            return 0;
        }
    }

    return 0;
}
\end{lstlisting}

\subsection*{Round~1088-F}
\begin{lstlisting}[style=cppcontest]
#include <bits/stdc++.h>
using namespace std;

const int P = 676767677;

long long pw(long long a, long long b) {
    long long r = 1;
    for (; b > 0; b >>= 1) {
        if (b & 1) r = r * a % P;
        a = a * a % P;
    }
    return r;
}

struct A {
    vector<long long> a, b;
    A(int n) : a(n+1), b(n+1) {
        a[0] = 1;
        for (int i = 1; i <= n; i++)
            a[i] = a[i-1] * i % P;
        b[n] = pw(a[n], P-2);
        for (int i = n; i >= 1; i--)
            b[i-1] = b[i] * i % P;
    }
    long long c(int n, int k) const {
        if (n < 0 || k < 0 || k > n) return 0;
        return a[n] * b[k] % P * b[n-k] % P;
    }
};

struct B {
    int a, b, c;
};

int main() {
    ios::sync_with_stdio(false);
    cin.tie(0);

    int t;
    cin >> t;
    vector<pair<int,int>> q(t);
    int mx = 0;
    for (int i = 0; i < t; i++) {
        cin >> q[i].first >> q[i].second;
        mx = max(mx, q[i].first + q[i].second);
    }

    A cb(mx);

    for (int i = 0; i < t; i++) {
        int n = q[i].first, m = q[i].second;
        long long base = cb.c(n+m-1, m-1);
        long long ans = 0;

        vector<B> st;
        st.push_back({1, n, 1});

        while (!st.empty()) {
            B cur = st.back();
            st.pop_back();

            int l = cur.a, r = cur.b, d = cur.c;
            int mid = (l + r) >> 1;
            int ls = mid - l;
            int rs = r - mid;
            int ss = r - l + 1;

            long long w = base;
            if (l > 1) w -= cb.c(n+m-ls-2, m-1);
            if (r < n) w -= cb.c(n+m-rs-2, m-1);
            if (l > 1 && r < n) w += cb.c(n+m-ss-2, m-1);
            w %= P;
            if (w < 0) w += P;

            ans = (ans + w * d) % P;

            if (l <= mid-1) st.push_back({l, mid-1, d+1});
            if (mid+1 <= r) st.push_back({mid+1, r, d+1});
        }

        cout << ans % P << '\n';
    }

    return 0;
}
\end{lstlisting}

\end{document}